\documentclass[12pt,a4paper]{article}
\usepackage[vietnamese,english]{babel}
\usepackage[T5,T1]{fontenc}
\newcommand*\en{\fontencoding{T1}\selectfont\selectlanguage{english}}
\newcommand*\vn{\fontencoding{T5}\selectfont\selectlanguage{vietnamese}}
\usepackage{amsmath, amssymb, fullpage, graphics}
\usepackage{caption, graphicx, fancyhdr, fancybox}
\usepackage{color}
\usepackage{tabularx}
\usepackage{longtable}
\usepackage{booktabs}
\usepackage{multicol}
\usepackage{multirow}
\usepackage{listings}
\usepackage{enumitem}
\usepackage{tvietlistings}
\usepackage[para]{footmisc} 
\usepackage{hyperref}
\usepackage{cite}

\usepackage{pgf}
\usepackage[numbers]{natbib}

\newcommand{\smallbacksquares}{\scalebox{0.6}{$\blacksquare$}}
\newcommand{\smallblacklozenge}{\scalebox{0.6}{$\blacklozenge $}}
\makeatletter
\fancypagestyle{footmark}{
  \ps@@fancy 
  \fancyfoot[C]{\footmark}
}

\makeatother
\begin{document}

\begin{center}
{\large\sf VietBinoculars: A Zero-Shot Approach for Detecting Vietnamese LLM-Generated Text}
\vspace{2mm}

Trieu Hai Nguyen$^{\text{a},\dagger}$, Sivaswamy Akilesh$^{\text{b},*}$ \\[2mm]

${^\text{a}}$ \textit{Faculty of Information Technology, Nha Trang University,\\
02 Nguyen Dinh Chieu Street, North Nha Trang 57134, Vietnam} \\
${^\text{b}}$ \textit{Faculty of Business Administration, Swiss School of Business and Management Geneva, 12 Avenue des Morgines, Geneva 1213, Switzerland} \\
[2mm] 
e-mails: $^\dagger$\textit{trieunh@ntu.edu.vn}; $^*$\textit{akilesh@ssbm.ch}

\end{center}
\begin{abstract}
The rapid proliferation of Large Language Models has intensified the challenge of distinguishing LLM-generated text from human writing in non-English languages. This study introduces VietBinoculars, a zero-shot detection framework coupling PhoGPT-4B observer and performer models with calibrated global decision thresholds. By utilizing specialized Vietnamese BPE tokenization, the method eliminates byte-level fragmentation and probability dilution common in massive multilingual backbones. Evaluated across multi-domain benchmarks, VietBinoculars achieves an area under the ROC curve exceeding 0.99. Under optimal Youden's J thresholds and greedy decoding, detection accuracy reaches at least 98.78\%, while significantly outperforming baseline Binoculars, zero-shot detectors, and commercial tools on creative Capybara prompts. Even under a strict false positive rate constraint of 0.06\%, the detector maintains F1-scores between 83.15\% and 94.70\%. Detection performance consistently improves with sequence length, stabilizing at optimal accuracy for passages containing 450 to 550 tokens. Extended stress testing across 48 distinct model-decoding configurations and three post-generation rewriting strategies delineates practical operational boundaries. VietBinoculars exhibits robust resilience against single-pass paraphrasing and human-style revisions, but experiences notable performance degradation under high-entropy sampling and iterative double paraphrasing.

\let\thefootnote\relax\footnotetext{\small $^{\dagger}$Corresponding author.
\\ 
Revised version submitted to VJCS
\\ 
\texttt{\small Preprint.} 
\hfill \textit{Oct 06, 2026} \\
}
\end{abstract}


\section{Introduction}
The text detection tasks performed by LLMs and other AI (Artificial Intelligence) systems play an important role in various real-world applications. LLM-generated text can be misused in various contexts, such as students using AI tools to complete assignments, or the automatic creation of misinformation and fake news on social media. These texts are often difficult for humans to distinguish from authentic content, potentially leading to misinformation and public misperception \cite{gehrmann-etal-2019-gltr}.  
In recent years, many studies have explored methods to detect if text is human-written or LLM-generated. In this context, Gehrmann et al. \cite{gehrmann-etal-2019-gltr} introduced GLTR (Giant Language Model Test Room) in 2019 as one of the first tools designed to detect LLM-generated texts. Based on statistical methods, GLTR calculates the ranking of the actual next word appearing in the input text compared to the model's prediction and visualizes it using colors. This laid the groundwork for subsequent zero-shot detection ideas, as the tool does not require to retrain any LLMs while still achieving competitive performance based on statistical features. However, due to its dependence on the model that generates the text, this approach faces challenges in detecting texts generated by commercial LLMs like ChatGPT\footnote{\url{https://chat.openai.com}}, Claude\footnote{\url{https://claude.ai}}, and its detection scope is quite limited.

One of the first large language models built on the GPT-2 architecture \cite{radford2019language} to generate propaganda news and fake articles was GROVER, proposed by Zellers et al. in 2019 \cite{zellers2019grover}. This model was evaluated to have higher reliability than human-written text. GROVER is particularly dangerous when employed for malicious purposes. In particular, the best way to detect texts generated by GROVER is to use GROVER itself, based on the exposure bias and applying variance reduction algorithms during text generation.

In 2020, Uchendu et al. \cite{uchendu-etal-2020-authorship} did not just address the challenge of distinguishing between LLM-generated and human-written text, but also formally defined the Authorship Attribution (AA) problem. This task involves identifying the specific Natural Language Generation (NLG) model responsible for producing a given text based on linguistic features such as n-grams, part-of-speech (POS) tags, topic modeling, LIWC, syntactic structure, or stylometric features. They also relied on classical classification methods such as Naive Bayes, SVM, Random Forest, KNN, as well as deep learning architectures like RNN, Stacked CNN, Parallel CNN, CNN-RNN. In 2021 and 2022, open-source autoregressive LLMs such as GPT-Neo, GPT-NeoX-20B \cite{gpt-neox}, and GPT-J-6B \cite{gpt-j} were released. These models have facilitated the generation of LLM-generated text samples, thereby enhancing the accuracy of detection methods, namely DetectGPT \cite{mitchell2023detectgpt} and Fast-DetectGPT \cite{fast-detectgpt}. Additionally, GPTZero\footnote{\url{https://gptzero.me}\label{fn:gptzero}} is a commercial application that was specifically developed to detect text generated by ChatGPT in 2023. Another example is Turnitin\footnote{\url{https://www.turnitin.com}\label{fn:turnitin}}, which is a well-known commercial platform for detecting LLM-generated text, but it particularly focuses on English, Spanish, and Japanese texts.

In the study of DetectGPT \cite{mitchell2023detectgpt}, Eric Mitchell et al. discovered that text samples generated by LLMs tend to lie in a region of negative curvature of the log-likelihood function of the model. This is also a zero-shot method, as it does not require retraining or collecting data for classification. Interestingly, compared to the above mentioned methods, DetectGPT analyzes the entire curvature of the log-likelihood probability function instead of relying on the average log-probability threshold per token. Although it is effective in detecting LLM-generated text, DetectGPT primarily results in high computational costs due to the need to create numerous perturbed variants of the input text by calling the model or API hundreds of times. An improvement over DetectGPT is Fast-DetectGPT \cite{fast-detectgpt}, which addresses its limitations by replacing the step of generating perturbed variants of the original text with a more efficient sampling step and uses conditional probability curvature to reduce the number of model calls.

Su et al. \cite{su2024log} proposed another zero-shot detection method for identifying LLM-generated text by analyzing the logarithmic rank of the next token predicted by LLMs. In 2024, Verma et al. introduced an advanced detector named Ghostbuster \cite{verma2024ghostbuster}. Ghostbuster is fine-tuned specifically to detect text generated by ChatGPT, but it is not effective in detecting outputs from other LLMs such as LLaMA \cite{llama3modelcard}. The principle of Ghostbuster allows human-written and LLM-generated texts to pass through weaker LLMs than the model being detected, in order to obtain the probability of the next token prediction, which serves as the main component of the extracted features. Finally, these features are passed through a basic classifier like Logistic Regression, to determine whether the text is human-written or LLM-generated. This method has the advantage of detecting black-box LLMs, such as ChatGPT, as it does not require access to next-token probability distributions from the model.

Building on zero-shot detection, Abhimanyu Hans et al. \cite{hans2024binoculars} proposed an additional detection model, referred to as Binoculars. The advantage of Binoculars is that it achieves state-of-the-art accuracy without requiring any training data at the time of its release, outperforming other zero-shot detection methods, including both commercial and open-source approaches. More importantly, Binoculars has practical significance in evaluating detection performance, achieving over 90\% accuracy in detecting English text generated by ChatGPT at a False Positive Rate (FPR) of 0.01\%, even though it was not trained on ChatGPT. Therefore, in this study, we employ Binoculars to detect Vietnamese text generated by LLMs such as Gemma 3-12B \cite{gemma3technicalreport}, Sailor2-8B \cite{sailor2sailingsoutheastasia}, and additional models. 
The main contributions of this study are summarized as follows:
\begin{itemize}
    \item An information-theoretic interpretation and schematic illustration are provided for cross-perplexity normalization in Binoculars, showing how cross-perplexity acts as an adaptive baseline that moderates score inflation on passages with low-probability vocabulary.
    
    \item VietBinoculars is formulated using a specialized Vietnamese model pair with a dedicated BPE tokenizer. An architectural ablation study against competitive multilingual models confirms that language-specific vocabularies prevent token fragmentation, while runtime profiling demonstrates a rapid, length-invariant median inference latency of 61.80 ms (mean 61.50 ms) on the evaluated GPU setup.
    
    \item We construct, validate, and publicly release standardized Vietnamese multi-domain benchmarks from journalism and literary works, establishing global decision thresholds under Youden's J, Closest Point, and a strict 0.06\% false positive rate criterion.
    
    \item Comprehensive out-of-domain evaluations demonstrate that VietBinoculars significantly outperforms baseline Binoculars, confirmed by paired McNemar contingency tests ($p < 0.001$), while creative Capybara evaluations demonstrate substantial advantages over commercial detectors.
    
    \item We conduct the first systematic robustness evaluation across 48 distinct model-decoding configurations (comprising 46 stochastic settings and two greedy controls) and three post-generation AI rewriting strategies. These experiments delineate empirical detection boundaries and identify multi-stage stochastic rewriting as the primary vulnerability for zero-shot detectors.
\end{itemize}

In Section~\ref{sec:VietBinoculars_overview} of this study, we provide an overview of the Binoculars method proposed by Hans et al.\cite{hans2024binoculars}. Then, we present the original Binoculars method adapted for detecting Vietnamese LLM-generated text, referred to as VietBinoculars.  Following this, in Section~\ref{sec:applicationbinocularstovietnamese}, we detail newly constructed Vietnamese LLM-generated datasets from the domains of news articles and literary works. Operating on a pair of LLMs, VietBinoculars computes Binoculars scores, which are then used to classify text through our globally optimized thresholds. These thresholds are specifically optimized for the detection of Vietnamese LLM-generated and human-written text. Based on our datasets, we evaluate the effectiveness of the VietBinoculars method under various metrics, including accuracy, F1-score, and AUC. We also compare VietBinoculars against the original Binoculars method, other state-of-the-art approaches, and commercial detection tools. Section~\ref{subsec:sampling_robustness} examines detector robustness across diverse decoding configurations, while Section~\ref{subsec:rewrite_robustness} evaluates resistance against three post-generation AI rewriting strategies. Section~\ref{sec:discussion} discusses challenges related to applying the VietBinoculars method to Vietnamese. Finally, Section~\ref{sec:conclusion} concludes the paper.

\section{Binoculars Method Overview}\label{sec:VietBinoculars_overview}
Currently, there are two main perspectives in academia on detecting whether text is human-written or LLM-generated. First, from the theoretical perspective of Varshney et al. (2020) \cite{varshney2020}, detecting LLM-generated text is considered impossible. However, from an empirical standpoint, any presence of advanced LLMs can be detected if a sufficiently large sample is available for statistical analysis or model training. To validate this perspective, two key issues need to be examined. First, we assess the reliability of the detector by detecting texts that have been intentionally modified to evade the detection system. Second, for practical applications, it is essential to carefully evaluate the effectiveness of the detector using the FPR metric. False positives are texts created by humans and yet labeled as LLM-generated. Only systems based on low FPR criteria are considered good and reliable for commercial applications while minimizing potential harm to users.

Binoculars follows the empirical perspective that it can detect text generated by modified prompts from any LLM at an extremely low FPR (0.01\%) without requiring retraining on those specific models. This method outperforms traditional approaches that use the same LLMs as a backbone for fine-tuning and retraining on samples for binary classification tasks \cite{verma2024ghostbuster}. However, Binoculars adopts a zero-shot approach, utilizing statistical features based on a globally optimal threshold to determine whether text is LLM-generated or human-written. The advantage of this method is that it does not require retraining on LLMs with hundreds of billions of parameters and numerous released versions while still being capable of detecting them.

In addition, Binoculars offers several advantages over other detectors, such as Ghostbuster, achieving state-of-the-art F1 scores (over 99\%) at the time of its release. According to the paper \cite{hans2024binoculars}, Binoculars demonstrates superior reliability compared to Ghostbuster on out-of-domain datasets. It exhibits high generalization capabilities, being able to detect texts generated by various LLMs such as LLaMa, Falcon, and ChatGPT without requiring detailed modifications for each model. In contrast, Ghostbuster is nearly ineffective at detecting texts generated by LLaMA. Importantly, Binoculars is unaffected by prompting strategies designed to bypass detection systems through prompt modification. Furthermore, Binoculars has commercial value due to its ability to detect black-box LLMs without requiring access to their next-token prediction probabilities unlike DetectGPT, Fast-DetectGPT, and GLTR, which rely on white-box LLMs. Finally, Binoculars performs well not only in English but also in various other languages, whereas well-known commercial tools, for instance Turnitin, largely operate in English. Moreover, it does not exhibit bias against texts generated by non-native English speakers in the task of determining whether English text is human-written or LLM-generated.

The Binoculars method uses a pair of closely related large language models sharing a pre-trained tokenizer. An observer model computes next-token predictions to establish a reference baseline for the input text. A performer model then calculates the perplexity of the actual input text to measure token surprise. In addition, the performer model evaluates its surprise regarding the predictions of the observer model. This evaluation yields the cross-perplexity metric. The Binoculars score is computed as the ratio of performer perplexity to cross-perplexity \cite{hans2024binoculars}. This score provides a robust criterion for distinguishing human-written text from LLM-generated text.

\subsection{VietBinoculars} 
According to the original Binoculars paper \cite{hans2024binoculars}, the authors primarily used a pair of Falcon-7B and Falcon-7B-Instruct models \cite{almazrouei2023falcon} to calculate Binoculars scores. However, the Falcon models are fundamentally trained on English and French. The authors noted that for raw text data in non-english languages, the vocabulary may not be well represented in the LLM. Languages such as Vietnamese typically exhibit lower True Positive Rates at low False Positive Rates and are more prone to mispredictions compared to languages that are well represented in the LLM. For our research, we used a pair of custom-designed PhoGPT models for Vietnamese representation, which have nearly 4 billion parameters and a context size of 8,192 \cite{nguyen2024phogpt}. To be specific, we use the PhoGPT-4B version as the observer model and PhoGPT-4B-Chat, a fine-tuned version of the original trained with 70K instructional prompts and 290K dialogue segments, as the performer model. PhoGPT is based on the Transformer decoder architecture for training and employs a BPE tokenizer tailored designed for Vietnamese \cite{vaswani2023attentionneed,trieu2022clustering}.

\begin{figure}[!ht]
\centering
\includegraphics[width=0.7\textwidth]{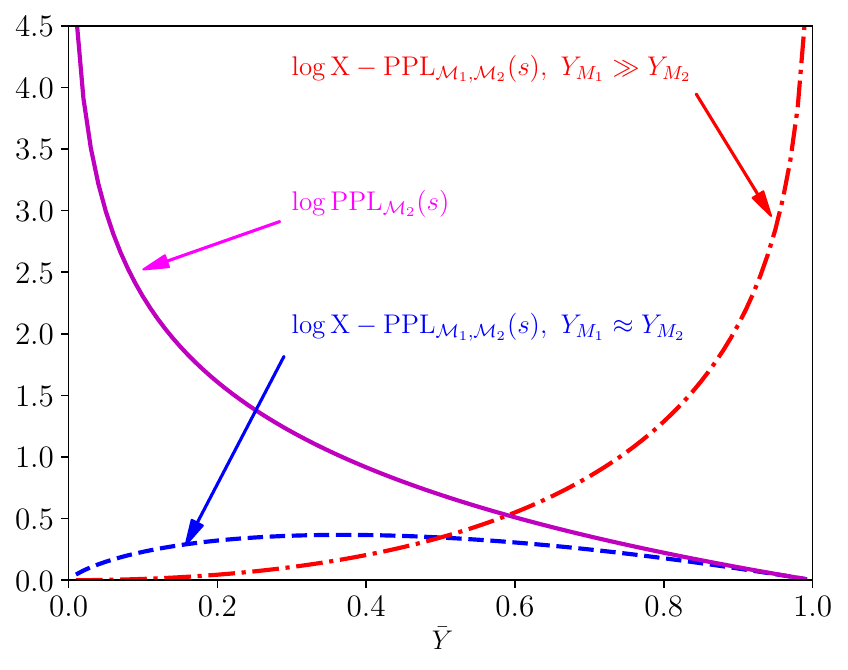}
\caption{Illustration of $\log{(\text{perplexity})}$ and $\log{(\text{cross-perplexity})}$. The x-axis represents the average next-token prediction probability $\bar{Y}$ for the same input string $s$. The solid magenta curve represents $\log{{\text{PPL}}_{\mathcal{M}_2}(s)}$ and the dashed curves represent $\log{{\text{X-PPL}}_{\mathcal{M}_1,\mathcal{M}_2}\left(s\right)}$ for different average next-token prediction probabilities of the observer and performer models, denoted $Y_{M_1}$ and $Y_{M_2}$, respectively. The dashed blue curve illustrates $\log{\text{X-PPL}_{\mathcal{M}_1,\mathcal{M}_2}(s)}$ when $Y_{M_1} \approx Y_{M_2}$, i.e., the observer and performer models are nearly identical. When the two models differ substantially, i.e., $Y_{M_1} \gg Y_{M_2}$, the cross-perplexity rises sharply, as depicted by the dash-dotted red curve.}
\label{fig:illustrationlogPPL_log_XPPL}
\end{figure}

In particular, for a Vietnamese input string $s$, the sequence is tokenized using the shared BPE tokenizer into $L$ tokens $\vec{x} = (x_1, \dots, x_L)$. Using PhoGPT-4B as the observer model $M_1$ and PhoGPT-4B-Chat as the performer model $M_2$, each model yields a probability distribution matrix $\mathbf{Y}^{(M_k)} = \mathcal{M}_k(s)$ of next-token predictions across each token position $i \in [1, L]$ over the shared vocabulary $\mathcal{V}$
\begin{equation}\label{eq:probabilitydistribution_concept}
  \mathbf{Y}^{(M_k)}_{ij} = \mathcal{M}_k(s)_{i,j} = P_{M_k}\left(v_j\middle| x_{<i}\right),\quad \forall v_j\in \mathcal{V},\ i=1:L,\ k\in\{1,2\},
\end{equation}
where $P_{M_k}\left(v_j\middle| x_{<i}\right)$ denotes the conditional probability assigned by model $M_k$ to the next token $v_j$ given the preceding context $x_{<i} = (x_1, \dots, x_{i-1})$. For the ground-truth token $x_i$ at each position, the performer model $M_2$ assigns probability $\mathcal{M}_2(s)_{i,x_i} = P_{M_2}(x_i \mid x_{<i})$. The average prediction probability across the sequence is denoted as $\bar{Y} = \frac{1}{L}\sum_{i=1}^L \mathcal{M}_2(s)_{i,x_i}$, which corresponds to the horizontal axis of Figure~\ref{fig:illustrationlogPPL_log_XPPL}, where $Y_{M_1}$ and $Y_{M_2}$ characterize the average next-token prediction probabilities of the observer and performer models, respectively.

The perplexity of the input string $s$ with respect to the performer model $M_2$ is defined as
\begin{equation}\label{eq:logppl_m2}
\log{{\text{PPL}}_{\mathcal{M}_2}\left(s\right)}=-\frac{1}{L}\sum_{i=1}^{L}{\log\left(\mathcal{M}_2\left(s\right)_{i,x_i}\right)},
\end{equation}
According to information theory \cite{shannon}, $\log{\text{PPL}_{\mathcal{M}_2}(s)}$ corresponds to the average cross-entropy of model $M_2$ over all tokens in $s$ (see~\ref{sec:appendixaveragecrossentropy}), given by 
\begin{equation}
  H(s, \mathcal{M}_2) = \log{{\text{PPL}}_{\mathcal{M}_2}\left(s\right)}.
\end{equation}
This quantity is depicted by the solid magenta curve in Figure~\ref{fig:illustrationlogPPL_log_XPPL}.
Next, using both probability distributions $\mathcal{M}_1(s)$ and $\mathcal{M}_2(s)$, cross-perplexity measures the expected surprise of the performer model $M_2$ with respect to the predictions of the observer model $M_1$.
\begin{equation}\label{eq:logxppl}
  \log{{\text{X-PPL}}_{\mathcal{M}_1,\mathcal{M}_2}\left(s\right)}=-\frac{1}{L}\sum_{i=1}^{L}{\mathcal{M}_1\left(s\right)_i}\cdot \log{\left(\mathcal{M}_2\left(s\right)_i\right)}.
\end{equation} 
Here, the dot product denotes the cross-entropy between the two next-token probability distribution vectors $\mathcal{M}_1(s)_i = \mathbf{Y}^{(M_1)}_i$ and $\mathcal{M}_2(s)_i = \mathbf{Y}^{(M_2)}_i$ at each token position \cite{hans2024binoculars,shannon}. The cross-perplexity is illustrated by the dashed curves in Figure~\ref{fig:illustrationlogPPL_log_XPPL}. When the two models are closely related ($Y_{M_1} \approx Y_{M_2}$), cross-perplexity closely tracks perplexity across different sequence difficulties, as shown by the dashed blue curve. This alignment provides an effective sequence-level baseline to normalize perplexity. When the two models differ substantially ($Y_{M_1} \gg Y_{M_2}$), cross-perplexity rises sharply, as depicted by the dash-dotted red curve.

The VietBinoculars score is calculated as the ratio of performer perplexity to cross-perplexity
\begin{equation}
B_{\mathcal{M}_1,\mathcal{M}_2}\left(s\right)=\frac{\log{\text{PPL}_{\mathcal{M}_2}\left(s\right)}}{\log{{\text{X-PPL}}_{\mathcal{M}_1,\mathcal{M}_2}\left(s\right)}}. \label{eq:binocularsscore}
\end{equation}
A higher VietBinoculars score indicates human-written text, whereas a lower score denotes LLM-generated text. This separation occurs because human-written text typically presents greater surprise to the performer model, resulting in a higher value of $\log{{\text{PPL}}_{\mathcal{M}_2}\left(s\right)}$. Meanwhile, the closely related model pair sharing the same tokenizer maintains a stable cross-perplexity baseline. Conversely, LLM-generated text yields a lower perplexity value due to model familiarity with generation patterns, thereby producing a smaller VietBinoculars ratio.

To illustrate why normalization by cross-perplexity is essential rather than relying solely on perplexity, consider an adversarial prompt such as ``Can you write a few sentences about a capybara being an astrophysicist?''. In this scenario, perplexity $\log{{\text{PPL}}_{\mathcal{M}_2}(s)}$ alone yields an unusually high score because tokens like ``capybara'' and ``astrophysicist'' are rare in common contexts. Solely relying on perplexity would cause standard detectors to misclassify this LLM-generated text as human-written. By contrast, cross-perplexity $\log{{\text{X-PPL}}_{\mathcal{M}_1,\mathcal{M}_2}(s)}$ also rises because the observer model finds the same unexpected tokens surprising. Taking the ratio between performer perplexity and cross-perplexity offsets this sequence-level inflation, preserving a low VietBinoculars score and correctly identifying the LLM-generated text (see Section~\ref{subsec:capybara_problem}).

Given a chosen classification threshold $\tau$, the binary decision rule $\hat{y} \in \{0, 1\}$ for classifying the input string $s$ is formally formulated as
\begin{equation}\label{eq:decision_rule}
\hat{y} = \begin{cases} 
1 \ (\text{LLM-generated text}), & \text{if } B_{\mathcal{M}_1,\mathcal{M}_2}(s) < \tau, \\ 
0 \ (\text{human-written text}), & \text{if } B_{\mathcal{M}_1,\mathcal{M}_2}(s) \ge \tau. 
\end{cases}
\end{equation}

\subsection{Evaluation Metrics for VietBinoculars} 
In essence, the task of detecting whether text is LLM-generated or human-written is formulated as a binary classification problem. We can use various binary classification evaluation metrics to assess the effectiveness of the VietBinoculars method, such as Accuracy, Precision, Recall, F1 Score, Confusion Matrix, ROC, or AUC. Most previous studies have employed AUC and F1 as evaluation metrics for detectors \cite{mitchell2023detectgpt,verma2024ghostbuster,hans2024binoculars}, whereas VietBinoculars introduces an additional practical evaluation metric that emphasizes the True Positive Rate at a very low False Positive Rate threshold (0.06\%) specifically for Vietnamese.
In this work, we define the negative class (labeled as 0) as human-written text, referring to content authored by humans, and the positive class (labeled as 1) as LLM-generated text. Based on the normalized Confusion Matrix, we have
\begin{equation}
\text{TPR}=\frac{TP}{TP+FN},\ \text{FPR}=\frac{FP}{FP+TN},
\end{equation}
in which $TP$ is the number of correctly detected LLM-generated texts; $FN$ is the number of LLM-generated texts incorrectly identified as human-written. $FP$ and $TN$ are the number of human-written texts incorrectly predicted as LLM-generated and the number of human-written texts correctly identified, respectively. We compare $\text{TPR}$ at a very low $\text{FPR}$ reference to minimize the potential harm that the detector may cause to users in practical applications. For example, if a student writes an essay but it is incorrectly labeled as LLM-generated, an error rate of less than or equal to 6 in 10,000 texts is considered acceptable for Vietnamese in practical use.

Based on the criteria mentioned above, we determine the global optimal threshold for classifying human-written and LLM-generated texts by analyzing the ROC curve using two widely adopted methods: Youden's J \cite{youden1950index} and the Closest Point approach \cite{PerkinsNeilJ.2006TIo}. Note that, instead of using the output as predicted class probabilities, we use the VietBinoculars score to represent points $(\text{FPR}, \text{TPR})$ on the graph when varying different thresholds to identify the global optimal threshold. The Youden's index determines the optimal threshold $t^\ast$ such that $J(t)$ attains its maximum value
\begin{equation}
J\left(t\right)=\text{TPR}\left(t\right) - \text{FPR}\left(t\right)
,\ t^\ast =  \underset{t}{\arg\max}  {\left(J\left(t\right)\right)}.
\end{equation}
The optimal threshold, according to Youden's index, is defined as the point representing the largest vertical distance from the random diagonal in the ROC graph. Next, the Closest Point method finds the point that is nearest to the optimal top-left corner, with coordinates (0,1), based on the Euclidean distance defined as follows
\begin{equation}
d\left(t\right)=\sqrt{\left(\text{FPR}\left(t\right)-0\right)^2+\left(1-\text{TPR}\left(t\right)\right)^2}
,\ t^\ast =  \underset{t}{\arg\min}  {\left(d\left(t\right)\right)}. \label{eq:closestpoint}
\end{equation}
The optimal threshold $t^\ast$ in Eq.~\ref{eq:closestpoint} represents the global optimal threshold determined by the Closest Point approach. This method prioritizes the selection of the optimal point that exhibits a low FPR and a high TPR.
In addition, we evaluate an operational cutoff constrained to $\text{FPR} \le 0.06\%$ on the calibration set to prioritize conservative false-alarm control in practical screening scenarios, defined as follows:
\begin{equation}
t^\ast =  \underset{\text{FPR}\left(t\right)\le0.06\%}{\arg\max}  {\left(\text{TPR}\left(t\right)\right)}.
\end{equation}

\section{Application of VietBinoculars to Vietnamese}\label{sec:applicationbinocularstovietnamese}
To apply the VietBinoculars method to Vietnamese, we first require datasets containing Vietnamese text samples that are either human-written or LLM-generated. However, most such datasets are currently unavailable. Thus, we construct datasets of Vietnamese human-written and LLM-generated text. We then apply the VietBinoculars method to the largest of these datasets to determine the optimal threshold. The detection performance of the method is subsequently evaluated on the remaining datasets, which serve as out-of-domain test sets. Moreover, we compare the performance of the VietBinoculars method with that of other methods on Vietnamese out-of-domain datasets.
\subsection{Vietnamese Datasets}\label{sec:vietnamese-dataset}
First, the Vietnamese texts written by humans are collected from the Vietnamese Online News Dataset on Kaggle \cite{nguyenhaitq} and from Vietnamese literary works authored by Vu Trong Phung\footnote{\url{https://doi.org/10.57967/hf/6218}}.  The Vietnamese Online News Dataset comprises news articles published up to July 2022 from reputable newspapers in Vietnam. The Vietnamese Literary Works Dataset, which contains both short stories and novels by the Vietnamese author Vu Trong Phung, has been segmented into smaller chunks of fewer than 512 tokens. In the text preprocessing stage for both datasets, we removed samples containing fewer than 50 words, as well as filtered out special characters, HTML tags, and links. Note that, for the VietBinoculars method to perform optimally, the length of news samples should be at least 50 tokens and should not exceed the context window size of the LLM model being used. Longer texts generally exhibit richer statistical features and are therefore easier to detect.
\begin{table}[!ht]
\centering
\caption{Summary and token-length distributions of Vietnamese datasets from two domains used in this study. The datasets are categorized into In-Domain and Out-of-Domain types. The number of documents (\#Docs) is evenly distributed between the Human-written and LLM-generated classes. Token counts are determined using the BPE tokenizer of the PhoGPT model, reporting mean, sample standard deviation (SD), median, and interquartile range [Q1, Q3] per class.} \label{table:dataset_detail} 
\resizebox{\textwidth}{!}{
\begin{tabular}{lllccccc}
\toprule
\multirow{2}{*}{\textbf{Domain}} & \multirow{2}{*}{\textbf{Dataset}} & \multirow{2}{*}{\textbf{Target}} & \multirow{2}{*}{\textbf{\# Docs}} & \multicolumn{2}{c}{\textbf{Human Tokens/Doc}} & \multicolumn{2}{c}{\textbf{LLM Tokens/Doc}} \\
\cmidrule(lr){5-6} \cmidrule(lr){7-8} 
& & & & \textbf{Mean $\pm$ SD} & \textbf{Median [IQR]} & \textbf{Mean $\pm$ SD} & \textbf{Median [IQR]} \\
\midrule
\multirow{3}{*}{News articles\textsuperscript{\ref{fn:news-dataset}}} 
& Sailor2-8B-OptiThreshold-News & In-Domain & 82,690 & 466.0 $\pm$ 142.1 & 467 [391, 524] & 575.8 $\pm$ 114.0 & 512 [512, 685] \\
& Sailor2-8B-Validation-News & In-Domain & 38,258 & 487.8 $\pm$ 159.1 & 469 [395, 594] & 627.4 $\pm$ 121.4 & 640 [528, 768] \\
& Gemma-3-12B-News & Out-of-Domain & 14,444 & 499.8 $\pm$ 163.4 & 483 [396, 611] & 592.0 $\pm$ 103.3 & 592 [488, 690] \\
\midrule
\multirow{2}{*}{Literary works\textsuperscript{\ref{fn:VTP-dataset}}} 
& Sailor2-8B-VuTrongPhung & Out-of-Domain & 738 & 478.0 $\pm$ 35.6 & 491 [462, 502] & 531.8 $\pm$ 14.3 & 532 [522, 541] \\
& Gemma-3-12B-VuTrongPhung & Out-of-Domain & 738 & 478.0 $\pm$ 35.6 & 491 [462, 502] & 575.3 $\pm$ 15.8 & 576 [566, 584] \\ 
\bottomrule
\end{tabular}
}
\end{table}

After collecting the human-written text samples, we use their first 50 tokens as prompts for LLMs to generate new text. By removing these initial 50 tokens from the generated output, we obtain the LLM-generated texts. The LLMs used in this study are Sailor2-8B-Chat \cite{sailor2sailingsoutheastasia} and Gemma-3-12B-it \cite{gemma3technicalreport}. Both are large, open-source language models, recently released in 2025, and capable of supporting multiple languages, including Vietnamese. Finally, we obtain the human-written and LLM-generated text datasets used in this study for VietBinoculars, as summarized in Table \ref{table:dataset_detail}. Specifically, we extracted and split the Vietnamese Online News Dataset into three subsets\footnote{\url{https://doi.org/10.57967/hf/6233} \label{fn:news-dataset}}: Sailor2-8B-OptiThreshold-News, Sailor2-8B-Validation-News, and Gemma-3-12B-News, corresponding to the Sailor2-8B-Chat and Gemma-3-12B-it models, respectively. Similarly, from the Vietnamese Literary Works Dataset, we generated two additional datasets\footnote{\url{https://doi.org/10.57967/hf/6234}\label{fn:VTP-dataset}}, Sailor2-8B-VuTrongPhung and Gemma-3-12B-VuTrongPhung, using the same procedure but with the respective LLMs. Remarkably, among multilingual LLMs with fewer than 10 billion parameters, Sailor2-8B demonstrates exceptional performance for Vietnamese. Therefore, the largest dataset Sailor2-8B-OptiThreshold-News is used to determine the optimal threshold for the VietBinoculars method in Vietnamese. For ease of understanding, Sailor2-8B-OptiThreshold-News can be regarded as the ``training'' dataset and Sailor2-8B-Validation-News as the ``validation'' dataset, both classified as In-Domain. The remaining datasets serve as ``test'' datasets and belong to the Out-of-Domain category. However, this analogy is only for illustrative purposes, as VietBinoculars does not require any training process.

As reported in Table~\ref{table:dataset_detail}, human news articles exhibit broader token-length dispersion with sample standard deviations between 142.1 and 163.4 tokens due to natural journalistic variability. In contrast, the literary datasets display tighter length distributions with standard deviations between 14.3 and 35.6 tokens because human excerpts were segmented into standardized chunks. Across all benchmarks, Table~\ref{table:dataset_detail} indicates class differences in token length: human medians range from 467 to 491 tokens, whereas LLM-generated medians range from 512 to 640 tokens. While all texts remain within the 768-token budget, these systematic differences warrant caution when interpreting classification margins and do not by themselves rule out length-related effects. Detailed dataset structures and per-class token-length distribution plots are provided in~\ref{sec:appendixdatasetillustration}.

\subsection{Experimental setup} \label{subsec:implementation_details}

For reproducibility, we report the key implementation parameters and hardware configuration. We use the official pre-trained checkpoints \texttt{vinai/PhoGPT-4B} as the observer model $M_1$ and \texttt{vinai/PhoGPT-4B-Chat} as the performer model $M_2$. Both models share the standard BPE tokenizer. Inference is executed in BFloat16 precision on an NVIDIA A100 Tensor Core GPU with 40GB VRAM. The host server runs Ubuntu 20.04 LTS equipped with an Intel Xeon Platinum 8259CL CPU and 32GB RAM. Input sequences are right-truncated to a maximum length of 768 tokens and padded to the longest sequence within each batch. Inference uses a batch size of 10 text pairs, evaluating 20 sequences per forward pass. On this setup, computing the VietBinoculars score requires an average latency of 61.50 ms per sample. This corresponds to an empirical throughput of 16.26 samples per second and 7,794 tokens per second across an average sequence length of 479 tokens. A detailed runtime profile across sequence lengths is provided in~\ref{sec:appendix_runtime_benchmark}.

\subsection{The Global Optimal Threshold for VietBinoculars} \label{subsec:findglobaloptimalthreshold} 
\begin{figure}[!htb]
  \centering
  \includegraphics[width=\textwidth]{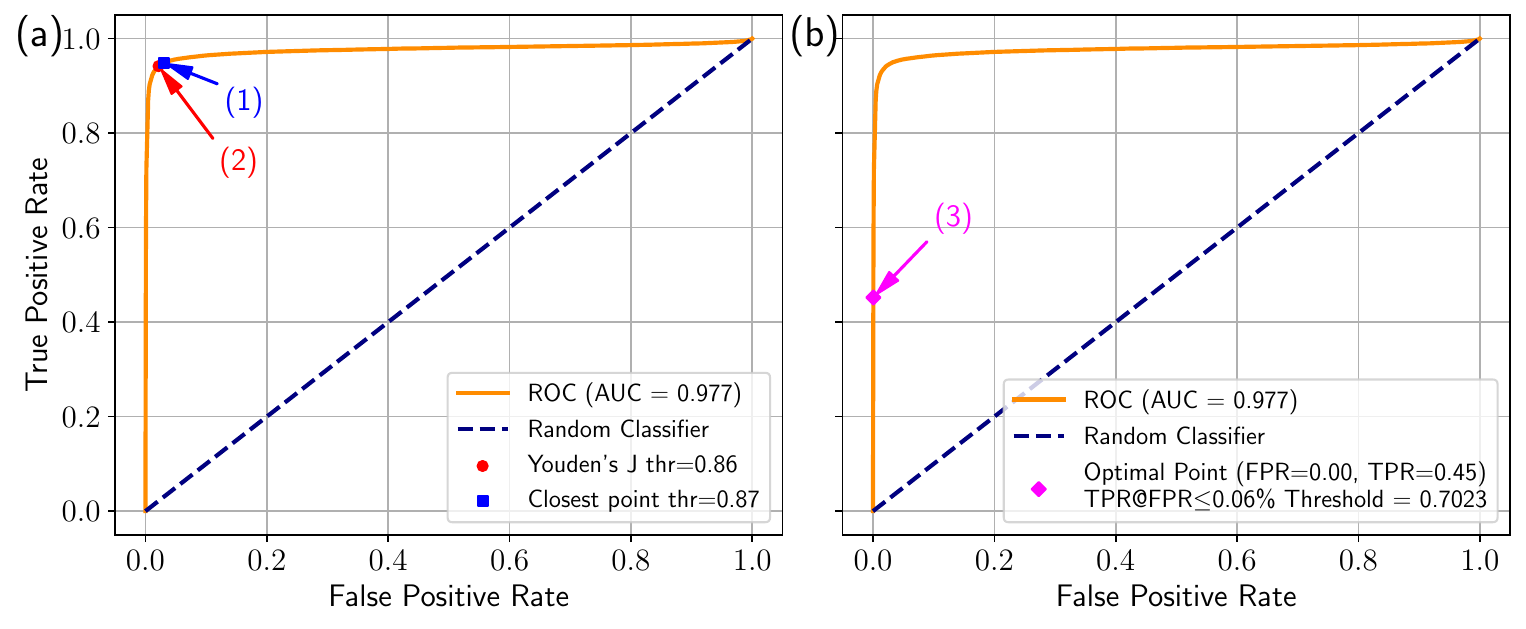}
  \caption{Global optimal threshold points represented on the ROC curve based on VietBinoculars scores for the Sailor2-8B-OptiThreshold-News dataset, showing (1) Youden's J point, (2) Closest point, and (3) optimal point at 0.06\% FPR. In subfigure (a), Youden's J and Closest thresholds are marked by a red circle [{\color{red}$\bullet$}] and a blue square [{\color{blue}$\smallbacksquares$}], respectively. The magenta diamond [{\color{magenta}$\smallblacklozenge$}] denotes the TPR@0.06\%FPR threshold in subfigure (b).}
  \label{fig:rocbinoculars}
\end{figure}

In this section, we calculate the VietBinoculars score on the largest dataset, Sailor2-8B-OptiThreshold-News, using the observer model $M_1$ and the performer model $M_2$. The VietBinoculars scores are computed using the formula in Eq.~\ref{eq:binocularsscore}. Based on these scores, we determine the global optimal threshold using Youden's J and Closest Point methods, as well as by selecting the global optimal point at 0.06\% FPR on the ROC curve, as shown in Figure~\ref{fig:rocbinoculars}. The optimal thresholds obtained were 0.86, 0.87, and 0.70 for Youden's J, Closest Point, and TPR@0.06\%FPR, respectively. We can observe that the TPR@0.06\%FPR threshold is lower than the other thresholds, since it must ensure that the FPR is approximately zero. 
Moreover, the Detection AUC of the VietBinoculars method on this dataset is approximately 0.98. With these thresholds, we can achieve a detection accuracy of 96\% for the Sailor2-8B-Validation-News dataset. The comparison of detection results, including Precision, Recall, Accuracy, and F1-score on the Sailor2-8B-OptiThreshold-News and Sailor2-8B-Validation-News datasets using the global thresholds identified above is presented in Table~\ref{tab:binoculars_optimal_thresholds}. 
\begin{table}[!htb]
\centering
\caption{Comparison of detection results using global thresholds of VietBinoculars on the Sailor2-8B-OptiThreshold-News and Sailor2-8B-Validation-News datasets. Results are reported for Youden's J, Closest Point, and TPR@0.06\%FPR thresholds. Zero-shot transfer yields consistent performance across both datasets, with both achieving the same highest F1-score of 0.96. Bold text denotes the highest F1-score within each column for the in-domain datasets.}
\label{tab:binoculars_optimal_thresholds} 
\resizebox{\textwidth}{!}{
\begin{tabular}{llccccccccccc}
\toprule
\multirow{2}{*}{\textbf{Dataset}} &  
  \textbf{Class/} &  
  \multicolumn{3}{c}{\textbf{Precision}} & 
  \multicolumn{3}{c}{\textbf{Recall}} & 
  \multicolumn{3}{c}{\textbf{F1-score}} & 
  \multirow{2}{*}{\textbf{Support}} \\ 
\cmidrule(lr){3-5} \cmidrule(lr){6-8} \cmidrule(lr){9-11}
& \textbf{Metric} & 
  \shortstack{Youden's\\J} & 
  \shortstack{Closest\\Point} & 
  \shortstack{TPR@\\0.06\%} & 
  \shortstack{Youden's\\J} & 
  \shortstack{Closest\\Point} & 
  \shortstack{TPR@\\0.06\%} & 
  \shortstack{Youden's\\J} & 
  \shortstack{Closest\\Point} & 
  \shortstack{TPR@\\0.06\%} & \\
\midrule
\multirow{5}{*}{\shortstack[l]{OptiThreshold\\News}} 
& Human        & 0.94 & 0.95 & 0.65 & 0.98 & 0.97 & 1.00 & 0.96 & 0.96 & 0.78 & 41345 \\
& LLM          & 0.98 & 0.97 & 1.00 & 0.94 & 0.95 & 0.45 & 0.96 & 0.96 & 0.62 & 41345 \\
& accuracy     & \multicolumn{6}{c}{}                    & 0.96 & 0.96 & 0.73 & 82690 \\
& macro avg    & 0.96 & 0.96 & 0.82 & 0.96 & 0.96 & 0.73 & \textbf{0.96} & \textbf{0.96} & 0.70 & 82690 \\
& weighted avg & 0.96 & 0.96 & 0.82 & 0.96 & 0.96 & 0.73 & 0.96 & 0.96 & 0.70 & 82690 \\
\midrule
\multirow{5}{*}{\shortstack[l]{Validation\\News}} 
& Human        & 0.95 & 0.96 & 0.66 & 0.98 & 0.97 & 1.00 & 0.96 & 0.96 & 0.79 & 19129 \\
& LLM          & 0.98 & 0.97 & 1.00 & 0.95 & 0.96 & 0.48 & 0.96 & 0.96 & 0.65 & 19129 \\
& accuracy     & \multicolumn{6}{c}{}                    & 0.96 & 0.96 & 0.74 & 38258 \\
& macro avg    & 0.96 & 0.96 & 0.83 & 0.96 & 0.96 & 0.74 & \textbf{0.96} & \textbf{0.96} & 0.72 & 38258 \\
& weighted avg & 0.96 & 0.96 & 0.83 & 0.96 & 0.96 & 0.74 & 0.96 & 0.96 & 0.72 & 38258 \\
\bottomrule
\end{tabular}%
}
\end{table}

\begin{figure}[!htb]
  \centering
  \includegraphics[width=\textwidth]{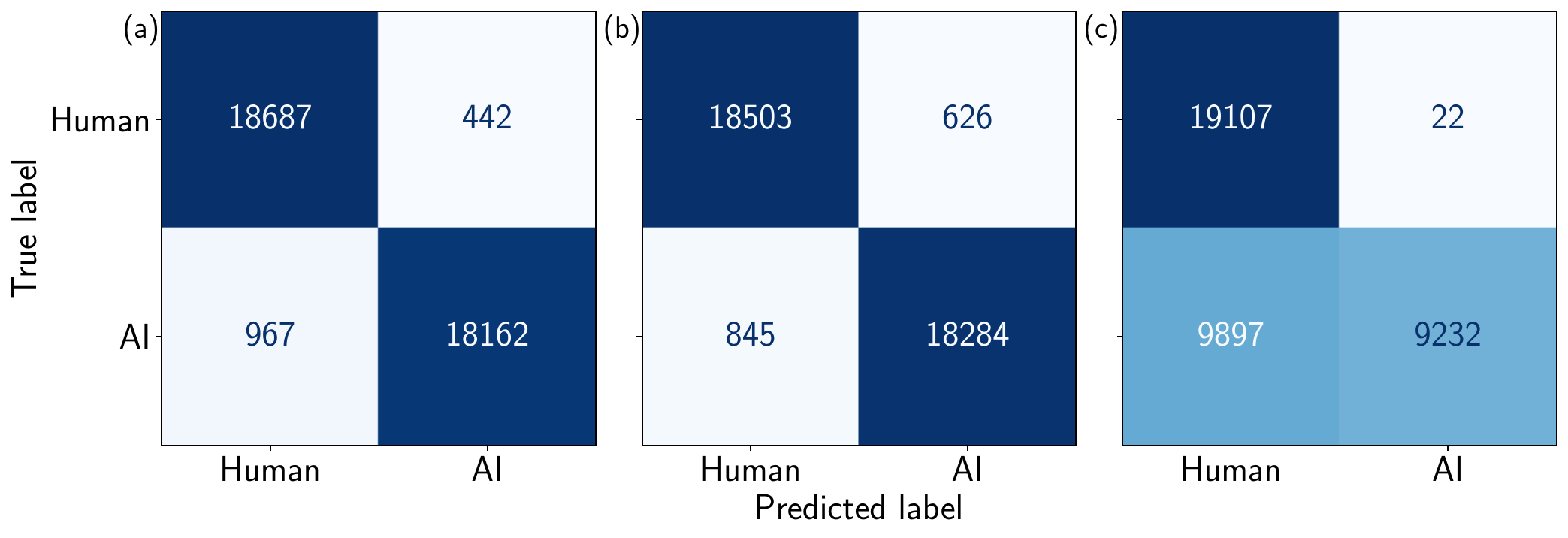}
  \caption{Confusion matrices of VietBinoculars on the Sailor2-8B-Validation-News dataset for (a) Youden's J threshold, (b) Closest Point threshold, and (c) TPR@0.06\%FPR threshold.}
  \label{fig:confusionmatrix_training}
\end{figure}

Overall, based on the results in Table~\ref{tab:binoculars_optimal_thresholds}, the Youden's J and Closest Point global thresholds yield a detection accuracy and F1-score of 96\%. The TPR@0.06\%FPR threshold achieves slightly lower performance, with 74\% accuracy and 72\% F1-score on the Sailor2-8B-Validation-News dataset. Specifically, the cutoff of 0.7023 was selected to satisfy an empirical FPR constraint of 0.06\% on the calibration dataset. On Sailor2-8B-Validation-News, it produces 22 false positives among 19,129 human-written documents, corresponding to an observed FPR of approximately 0.115\%. This empirical discrepancy demonstrates that the calibration target does not guarantee an identical false positive rate on unseen data. Nevertheless, this conservative operating point substantially mitigates false accusations compared to unconstrained thresholds. Additionally, detailed detection results on the Sailor2-8B-Validation-News dataset are presented in the Confusion Matrix shown in Figure~\ref{fig:confusionmatrix_training}.

\subsection{Applying VietBinoculars to Out-of-Domain Datasets} 
To examine the generalization capability of VietBinoculars, we evaluate its detection performance on out-of-domain benchmarks using the global thresholds established in Section~\ref{subsec:findglobaloptimalthreshold}. Specifically, we evaluate Gemma-3-12B-News from the news domain, together with Gemma-3-12B-VuTrongPhung and Sailor2-8B-VuTrongPhung from the literary domain. Table~\ref{tab:vietbioculars_on_out_of_domain} presents the empirical results across all thresholds, including raw confusion counts, 95\% Wilson score confidence intervals for accuracy, and paired McNemar significance tests against the original Binoculars baseline.

\begin{table}[!htb]
\centering
\small
\caption{Generalization performance of VietBinoculars across out-of-domain benchmarks, including raw confusion counts, 95\% Wilson score confidence intervals for accuracy, and paired McNemar significance tests against original Binoculars. Global decision thresholds determined on the training set are Youden's J ($\tau = 0.8607$), Closest Point ($\tau = 0.8729$), and TPR@0.06\%FPR ($\tau = 0.7023$). Macro F1-score is reported across Human and AI classes in alignment with Table~\ref{tab:binoculars_optimal_thresholds}. Paired McNemar tests employ Edwards' continuity correction against baseline Binoculars at its standard threshold ($\tau = 0.9015$).}
\label{tab:vietbioculars_on_out_of_domain} 
\resizebox{\textwidth}{!}{
\begin{tabular}{llccccccc}
\toprule
\textbf{Dataset} & \textbf{Threshold} & \textbf{TP} & \textbf{FP} & \textbf{TN} & \textbf{FN} & \shortstack{\textbf{Accuracy}\\\textbf{(95\% Wilson CI)}} & \shortstack{\textbf{Macro}\\\textbf{F1-score}} & \shortstack{\textbf{McNemar}\\\textbf{$p$-value}} \\
\midrule
\multirow{3}{*}{\shortstack[l]{Sailor2-8B-\\VuTrongPhung\\($N = 738$)}} 
& Youden's J & 369 & 0 & 369 & 0 & \textbf{1.0000} [0.9948, 1.0000] & \textbf{1.0000} & $< 0.001$ \\
& Closest Point & 369 & 0 & 369 & 0 & \textbf{1.0000} [0.9948, 1.0000] & \textbf{1.0000} & $< 0.001$ \\
& TPR@0.06\%FPR & 248 & 0 & 369 & 121 & 0.8360 [0.8076, 0.8610] & 0.8315 & $< 0.001$ \\
\midrule
\multirow{3}{*}{\shortstack[l]{Gemma-3-12B-\\VuTrongPhung\\($N = 738$)}} 
& Youden's J & 369 & 1 & 368 & 0 & \textbf{0.9986} [0.9924, 0.9998] & \textbf{0.9986} & $< 0.001$ \\
& Closest Point & 369 & 1 & 368 & 0 & \textbf{0.9986} [0.9924, 0.9998] & \textbf{0.9986} & $< 0.001$ \\
& TPR@0.06\%FPR & 330 & 0 & 369 & 39 & 0.9472 [0.9286, 0.9611] & 0.9470 & $< 0.001$ \\
\midrule
\multirow{3}{*}{\shortstack[l]{Gemma-3-12B-\\News\\($N = 14,444$)}} 
& Youden's J & 7,207 & 161 & 7,061 & 15 & \textbf{0.9878} [0.9859, 0.9895] & \textbf{0.9878} & $< 0.001$ \\
& Closest Point & 7,211 & 229 & 6,993 & 11 & 0.9834 [0.9812, 0.9853] & 0.9834 & $< 0.001$ \\
& TPR@0.06\%FPR & 6,298 & 3 & 7,219 & 924 & 0.9358 [0.9317, 0.9397] & 0.9356 & $< 0.001$ \\
\bottomrule
\end{tabular}}
\end{table}

The empirical findings in Table~\ref{tab:vietbioculars_on_out_of_domain} demonstrate remarkable generalization across out-of-domain benchmarks. Under Youden's J and Closest Point thresholds, VietBinoculars achieves perfect discrimination on Sailor2-8B-VuTrongPhung with zero false alarms and zero misses. On Gemma-3-12B-VuTrongPhung, the detector misclassifies only a single document across 738 samples, achieving an accuracy and F1-score of 0.9986 with a narrow confidence interval of $[0.9924, 0.9998]$. This confirms that the near-perfect scores on literary texts originate from distinct linguistic separability rather than rounding artifacts. Under the conservative TPR@0.06\%FPR threshold, VietBinoculars produces zero observed false positives on the two evaluated literary benchmarks while preserving substantial AI detection recall. Finally, paired McNemar contingency tests demonstrate that the performance advantage of VietBinoculars over the original Binoculars method is statistically significant across all benchmarks and thresholds ($p < 0.001$).  
\begin{figure}[!htb]
  \centering
    \includegraphics[width=\textwidth]{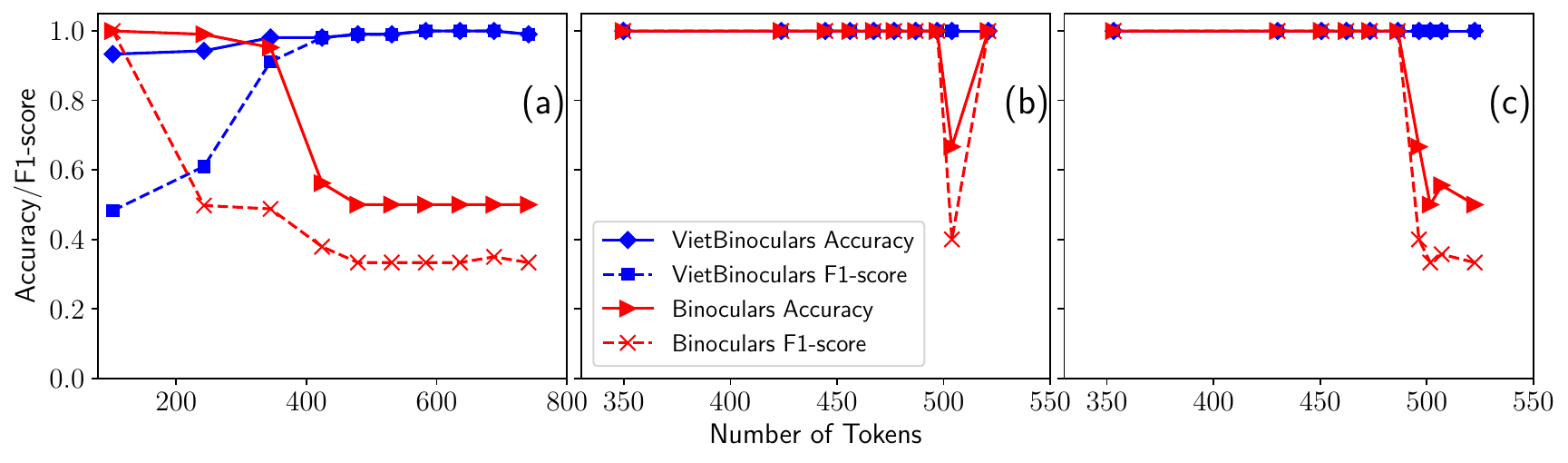}
  \caption{Detection performance versus BPE token length across out-of-domain benchmarks: (a) Gemma-3-12B-News, (b) Gemma-3-12B-VuTrongPhung, and (c) Sailor2-8B-VuTrongPhung. Solid and dashed curves denote Accuracy and macro F1-score, respectively. Markers show original Binoculars ({\color{red}$\blacktriangleright$}, {\color{red}$\times$}) and VietBinoculars under Youden's J threshold ({\color{blue}$\blacklozenge$}, {\color{blue}$\blacksquare$}). Curves are computed across ten equal-frequency decile bins on length-aligned subsets: $N = 1,046$ (~105 docs/bin) for news in (a), $N = 77$ (7--8 docs/bin) in (b), and $N = 90$ (9 docs/bin) in (c). Axis spans reflect empirical token support (80--800 tokens in news; 330--550 in literature).}
  \label{fig:impact_of_text_length}
\end{figure}

In the subsequent step, we study the impact of the number of tokens on the detection performance of VietBinoculars. We conduct experiments on the out-of-domain datasets and compare the results with the original Binoculars, which demonstrates good performance primarily in English. We use Accuracy and F1-score, represented by solid and dashed curves respectively, to compare detection performance. Figure~\ref{fig:impact_of_text_length} shows that the original Binoculars method performs worse than VietBinoculars on Vietnamese text, particularly when the number of tokens is sufficiently large. Moreover, the results presented in Fig.~\ref{fig:impact_of_text_length}(a) indicate that as sequence length increases across decile bins, the detection performance of VietBinoculars improves significantly on the Gemma-3-12B-News dataset. Similarly, the Binoculars method shows a notable decline in performance once the input length exceeds 480 tokens on the Gemma-3-12B-VuTrongPhung and Sailor2-8B-VuTrongPhung datasets (see Fig.~\ref{fig:impact_of_text_length}(b), (c)). Clearly, longer text samples provide richer statistical features, making them easier to detect using VietBinoculars or other zero-shot detection methods. 

\begin{figure}[!htb]
  \centering
    \includegraphics[width=\textwidth]{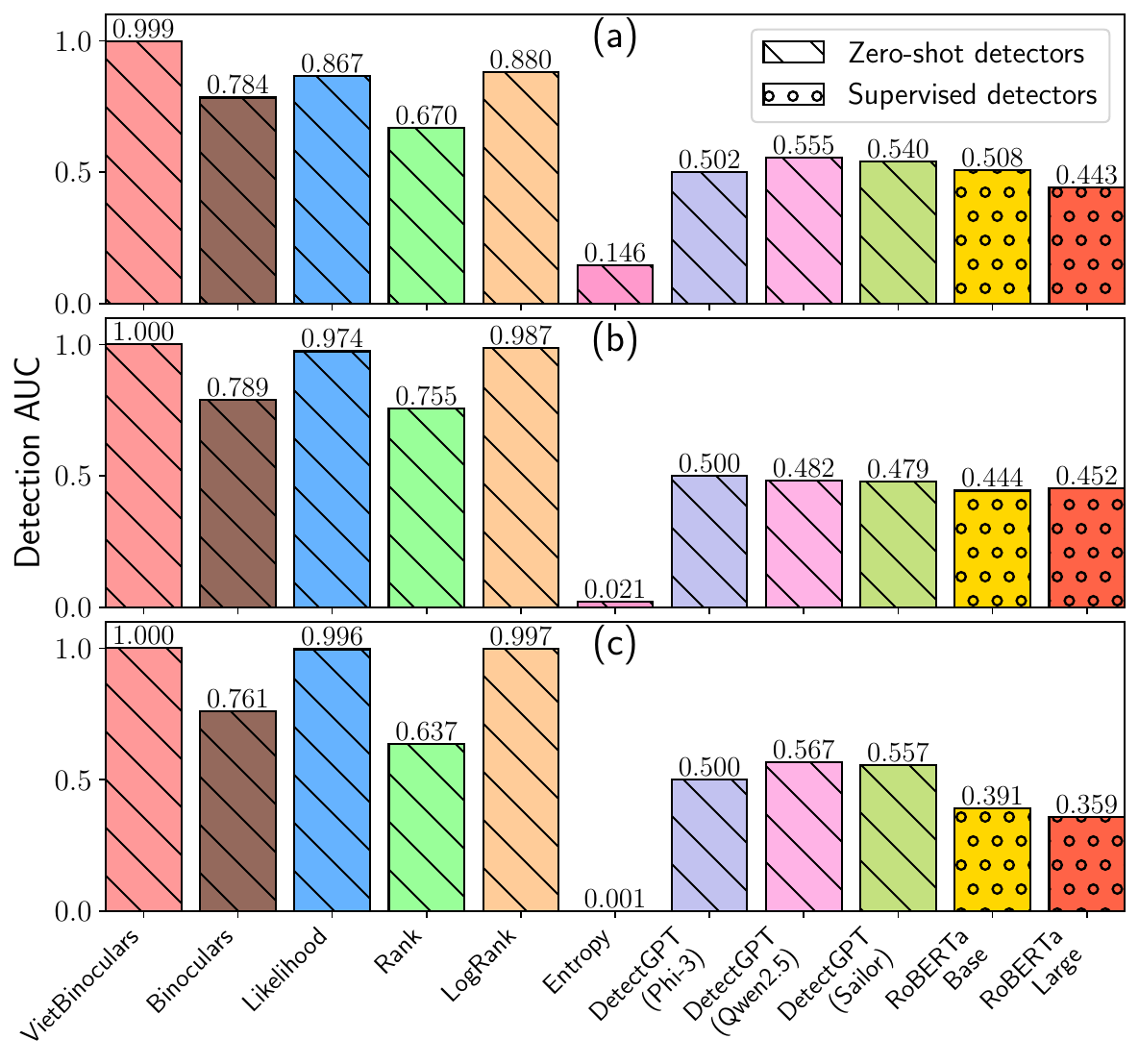}
  \caption{Detection AUROC for various zero-shot and supervised learning methods on Vietnamese out-of-domain datasets: (a)--Gemma-3-12B-News; (b)--Gemma-3-12B-VuTrongPhung; (c)--Sailor2-8B-VuTrongPhung. Zero-shot detectors are represented with the ``\texttt{\textbackslash}'' hatch pattern, while supervised learning detectors are represented with the ``o'' hatch pattern. DetectGPT uses top-$k$ and top-$p$ sampling with parameters $k=40$ and $p=0.96$.}
  \label{fig:detection_auc_comparison}
\end{figure}
Furthermore, the comparison of Detection AUC between zero-shot methods and supervised learning methods on Vietnamese out-of-domain datasets is presented in Figure~\ref{fig:detection_auc_comparison}. The zero-shot methods include VietBinoculars, Binoculars \cite{hans2024binoculars}, Likelihood (average per-token log probability), Rank/LogRank (average observed rank or log-rank of tokens), Entropy (average entropy in the model's predictive distribution) and DetectGPT \cite{mitchell2023detectgpt,gehrmann-etal-2019-gltr}. The supervised learning methods include OpenAI's RoBERTa-based GPT-2 detector models \cite{solaiman2019releasestrategiessocialimpacts,liu2019robertarobustlyoptimizedbert}. The results in Figure~\ref{fig:detection_auc_comparison} show that VietBinoculars consistently outperforms the other zero-shot methods across all datasets. Notably, VietBinoculars achieves a Detection AUC greater than 0.99, which is significantly higher than the approximately 0.76 obtained by the original Binoculars method on Vietnamese out-of-domain datasets. 
Because AUROC evaluates classification performance across all possible cutoffs, it decouples model representation from threshold calibration. This substantial margin demonstrates that empirical gains originate directly from Vietnamese-centric language models rather than threshold tuning. 
Likewise, VietBinoculars surpasses the performance of other zero-shot methods, such as Rank, Entropy and several variants of DetectGPT. In this experiment, we use FLAN-T5 Large \cite{chung2022scalinginstructionfinetunedlanguagemodels}, a pre-trained mask-filling model with support for Vietnamese, to generate 20 perturbed variants of each input for the DetectGPT method. Specifically, we randomly mask 2-word spans until 15\% of the words are replaced. Sailor-7B\footnote{\url{https://huggingface.co/sail/Sailor-7B}}, Qwen2.5-7B\footnote{\url{https://huggingface.co/Qwen/Qwen2.5-7B}} and Phi-3-mini-4k-instruct\footnote{\sloppy \url{https://huggingface.co/microsoft/Phi-3-mini-4k-instruct}} are used as source models for detection evaluation. On the two literary benchmarks, VietBinoculars achieves an AUROC of 1.000 to three decimal places, substantially exceeding the OpenAI RoBERTa detectors, whose displayed AUROCs range from 0.359 to 0.452. The failure of the OpenAI detector models can be explained by the fact that they were trained only to detect text generated by GPT-2, whereas our out-of-domain dataset was produced by recently released state-of-the-art LLMs.

The poor performance of competing detectors stems from four primary factors. First, language coverage remains severely limited in English-centric models. Baselines such as Falcon-7B and OpenAI RoBERTa were pre-trained predominantly on English corpora, preventing them from capturing Vietnamese syntactic structures. Second, English-oriented tokenizers split accented Vietnamese syllables into multiple byte-level tokens. This artificial fragmentation distorts transition probabilities and inflates sequence perplexity. In contrast, the dedicated 20,480-token BPE vocabulary of PhoGPT-4B preserves complete syllables and yields reliable token likelihoods. Third, multilingual backbones suffer from probability dilution. In vocabularies exceeding 151,665 tokens, the softmax denominator disperses probability mass across many irrelevant languages. As shown in our ablation study in~\ref{sec:appendix_model_pair_ablation} and Table~\ref{tab:model_pair_ablation}, this dilution increases entropy noise and lowers discriminative sensitivity. Finally, threshold selection compounds detector failure in commercial tools. Classifiers such as GPTZero and Radar apply a static 0.5 probability cutoff, which becomes severely uncalibrated on non-English text. Meanwhile, the low AUROC of zero-shot baselines in Figure~\ref{fig:detection_auc_comparison} proves that their deficiency is intrinsic to representation rather than decision boundary selection.

These results establish strong discrimination on the evaluated out-of-domain benchmarks. Their extension to sampled generations is examined in Section~\ref{subsec:sampling_robustness}, where performance varies with both the generation model and the decoding parameters.

\subsection{Capybara Problem in Vietnamese} \label{subsec:capybara_problem} 
In this section, we evaluate the effectiveness of the VietBinoculars method in a special case where users intentionally prompt LLMs to generate text, for example: ``{\vn \textit{Bạn có thể viết một vài câu kể chuyện về một con capybara là một nhà vật lý thiên văn không?}}'' (English version is ``\textit{Can you write a few sentences telling a story about a capybara being an astrophysicist?}''). The LLMs\footnote{\url{https://poe.com}} used to generate text from the capybara prompt include GPT-4o, GPT-4.1, GPT-4.1-mini, GPT-3.5-Turbo, Llama-3.1-405B, Llama-4-Scout-Nitro, Llama-4-Maverick-T, Grok-3, Gemini-2.5-Flash, Gemini-2.5-Pro, Gemma-3-27B, Claude-3-Sonnet, Claude-3-Haiku, Claude-3.5-Sonnet, Claude-3.7-Sonnet, DeepSeek-R1, DeepSeek-V3, Qwen3-235B-A22B-FW, Qwen-2.5-72B-T, Mistral-Medium, and Mistral-Large-2. In this case, detection models based on statistical features tend to misclassify significantly due to the unexpected presence of tokens such as ``\textit{capybara}'' and ``{\vn \textit{nhà vật lý thiên văn}}'' (English version is ``\textit{astrophysicist}'') . 
\begin{figure}[!htbp]
  \centering
  \includegraphics[width=\textwidth]{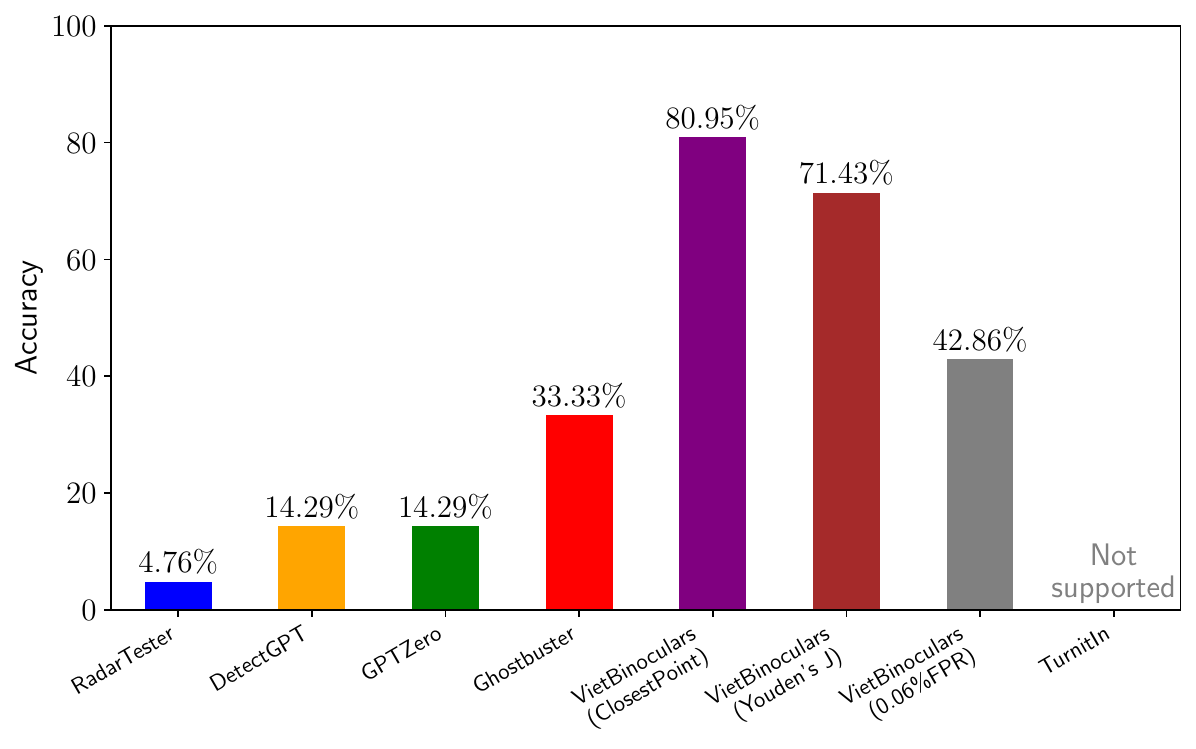}
  \caption{Comparison of VietBinoculars with popular detection methods on the Capybara problem in Vietnamese Language. The x-axis denotes different detection methods, while the y-axis indicates their detection accuracy. VietBinoculars was evaluated using three different global thresholds as described above. Other detection methods, such as RadarTester, which is based on four backbone models (Dolly V2-3B, Camel-5B, Dolly V1-6B, and Vicuna-7B), along with DetectGPT, GPTZero, and Ghostbuster, all apply a default prediction threshold of 0.5 for classification. Commercial tools exhibit poor detection performance on the Vietnamese Capybara problem.} 
  \label{fig:genneral_capybara_detection_results}
\end{figure}
Based on the optimal threshold for Vietnamese determined in Section~\ref{subsec:findglobaloptimalthreshold}, we compare the detection performance of VietBinoculars with other popular detection methods, including RadarTester \cite{hu2023radar}, the commercial products DetectGPT\footnote{\url{https://detectgpt.com}} and GPTZero\textsuperscript{\ref{fn:gptzero}}, as well as Ghostbuster \cite{verma2024ghostbuster}. All commercial tools, including DetectGPT, GPTZero and Turnitin, were accessed in June 2025 via their official Web interfaces using default settings without modification. In order to determine whether a text is more likely LLM-generated or human-written using RadarTester, we rely on four backbone models, namely Dolly V2-3B, Camel-5B, Dolly V1-6B, and Vicuna-7B. The closer the value is to 1, the more likely the text is classified as LLM-generated; conversely, the closer it is to 0, the more likely it is classified as human-written. In this experiment, the default prediction threshold for probabilities is typically set at 0.5. If any of the four models produces a value greater than 0.5, the text is classified as LLM-generated. However, RadarTester is only effective in detecting capybara texts when they are generated by DeepSeek-R1, while failing to detect those generated by other LLMs (see \ref{subsec:appendixcapybaradetectionresults}). Similarly, the commercial detection tools and Ghostbuster also adopt default thresholds of 0.5, classifying a text as LLM-generated when the predicted probability is greater than or equal to 0.5. In fact, the default thresholds used by these tools are not really reliable to classify texts generated by LLMs or written by humans. 
Figure~\ref{fig:genneral_capybara_detection_results} presents the detection results of VietBinoculars with different optimal thresholds, along with other popular detectors, on the Capybara problem in Vietnamese.  We observe that the Binoculars detection model demonstrates superior performance, achieving 80.95\% accuracy when using the optimal Closest Point threshold. Even the Binoculars detection model with TPR@0.06\%FPR reliability achieved higher detection accuracy (42.86\%) than Ghostbuster (33.33\%), which recorded the highest accuracy among the comparison tools mentioned above. Meanwhile, the commercial tools GPTZero and DetectGPT only achieve 14.29\% accuracy, while Turnitin does not even support Vietnamese. 

\subsection{Robustness to Decoding Parameters}\label{subsec:sampling_robustness}

The earlier benchmarks rely on greedy decoding, and their high detection scores may depend on this procedure. We therefore extend the main evaluation to sampled text. This experiment examines a scenario where decoding parameters are varied while keeping both detectors fixed. It investigates a potential avenue for reducing detectability without evaluating adaptive attacks or post-generation modifications.

The 369 human-written passages from the Vu Trong Phung literary dataset serve as source prompts for three generation models, including Sailor2-8B-Chat, Sailor2-14B-Chat~\cite{sailor2sailingsoutheastasia} and Gemma-3-12B-it. For each source passage and configuration, each model generates a single continuation. The parameter grid is defined as
\begin{equation}
\mathcal{G}=\{0.7,1.0,1.2,1.5\}\times\{0.65,0.80,0.95,1.0\},
\end{equation}
where $T$ denotes temperature and $p$ denotes the nucleus sampling parameter (top-$p$). This yields 16 configurations per model, each comprising 369 human-written and 369 LLM-generated texts. Across all sampled configurations, prompts remain identical for each source passage. The $(T,p)=(1.0,1.0)$ settings for Sailor2-8B-Chat and Gemma-3-12B-it serve as deterministic greedy controls, whereas the corresponding Sailor2-14B-Chat configuration uses sampling. Because human passages are shared across configurations, the resulting 48 evaluation sets\footnote{Detailed records and text samples for all 48 configurations are available at \url{https://doi.org/10.57967/hf/10537}.} are not statistically independent collections of human text.

\begin{figure}[!htb]
\centering
\includegraphics[width=\textwidth]{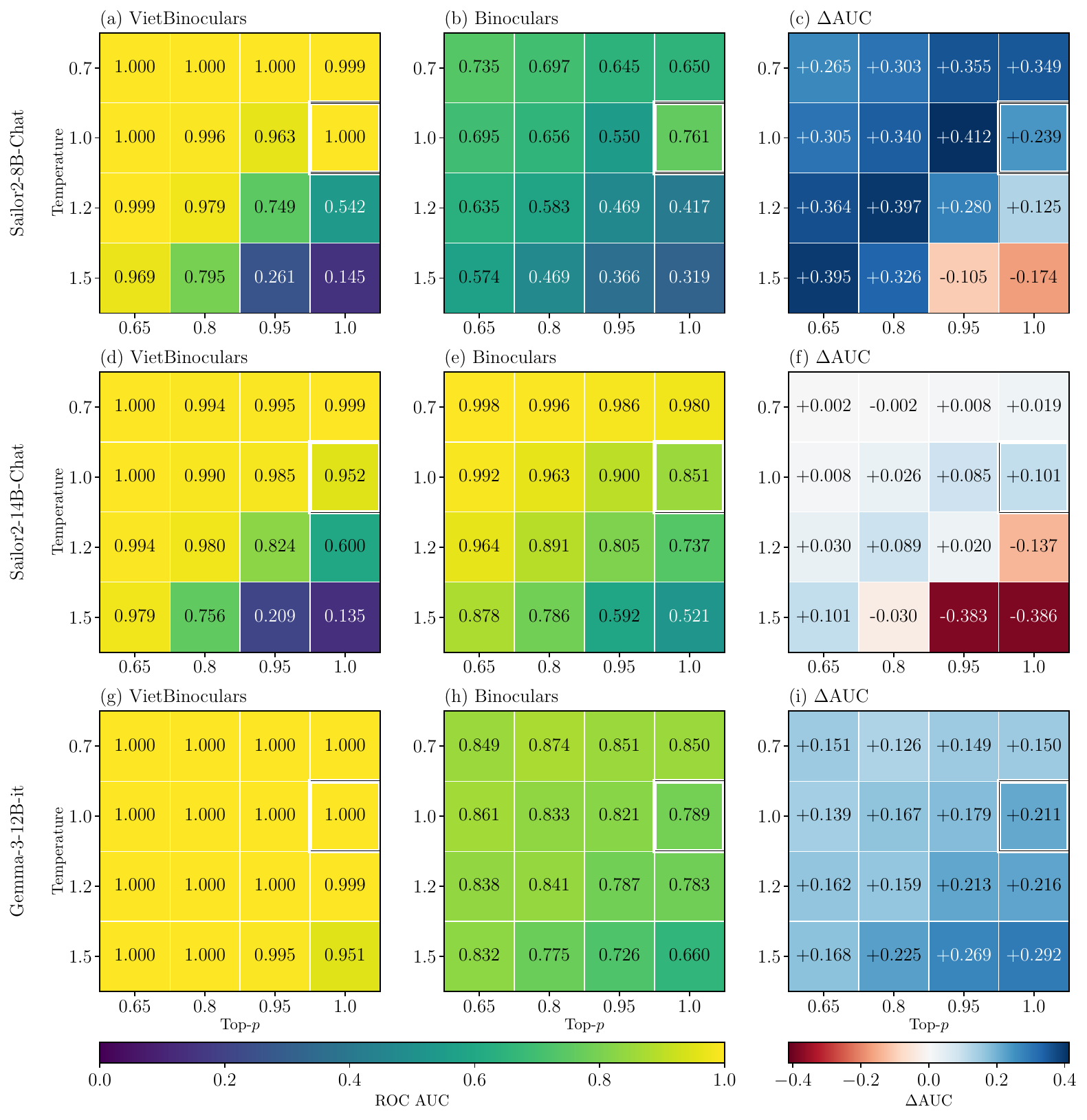}
\caption{ROC AUC across varying temperature and top-$p$ values. Rows correspond to Sailor2-8B-Chat, Sailor2-14B-Chat, and Gemma-3-12B-it. Columns display VietBinoculars, Binoculars, and their AUC differences, where positive values indicate higher AUC for VietBinoculars. Absolute AUC panels share a common color scale, whereas difference panels use a symmetric scale centered at zero. Outlined $(1,1)$ cells denote greedy controls for Sailor2-8B-Chat and Gemma-3-12B-it, while the corresponding Sailor2-14B-Chat cell represents a sampled configuration.}
\label{fig:sampling_robustness}
\end{figure}

We evaluate detection performance using ROC AUC and binary F1 on the 48 evaluation sets. Both detectors apply their previously established decision thresholds without recalibration on this grid, corresponding to Youden's J threshold for VietBinoculars and $0.902$ for Binoculars. Across all configurations, we report both the mean performance and the minimum observed scores in Table~\ref{table:sampling_robustness} to assess worst-case stability. 
\begin{table}[!htb]
\centering
\caption{Detection performance of VietBinoculars and Binoculars across 16 decoding configurations for each generation model. The reference $(T,p)=(1.0,1.0)$ setting employs greedy decoding for Sailor2-8B and Gemma-3-12B, whereas Sailor2-14B uses sampling. All reference configurations are included in the reported grid means. Binary F1 is computed using the fixed decision thresholds described in the text, where higher values indicate superior performance.}
\label{table:sampling_robustness}
\resizebox{\textwidth}{!}{%
\begin{tabular}{llccccc}
\toprule
Generation model & Detector & \shortstack{AUC\\$(T,p)=(1,1)$} & \shortstack{Mean\\AUC} & \shortstack{Minimum\\AUC} & \shortstack{Mean\\F1} & \shortstack{Minimum\\F1} \\
\midrule
Sailor2-8B & VietBinoculars & 1.000 & 0.837 & 0.145 & 0.587 & 0.000 \\
 & Binoculars & 0.761 & 0.576 & 0.319 & 0.001 & 0.000 \\
\midrule
Sailor2-14B & VietBinoculars & 0.952 & 0.837 & 0.135 & 0.626 & 0.005 \\
 & Binoculars & 0.851 & 0.865 & 0.521 & 0.220 & 0.000 \\
\midrule
Gemma-3-12B & VietBinoculars & 1.000 & 0.997 & 0.951 & 0.962 & 0.570 \\
 & Binoculars & 0.789 & 0.811 & 0.660 & 0.001 & 0.000 \\
\bottomrule
\end{tabular}%
}
\end{table}
Table~\ref{table:sampling_robustness} shows that the mean AUC of VietBinoculars decreases by 16.3, 11.5, and 0.3 percentage points compared to the reference $(T,p)=(1.0,1.0)$ configuration for Sailor2-8B-Chat, Sailor2-14B-Chat, and Gemma-3-12B-it, respectively. Despite these reductions, VietBinoculars outperforms the original Binoculars method in most cases. The only exception occurs on Sailor2-14B-Chat, where Binoculars achieves a higher mean AUC by 2.8 percentage points. Furthermore, VietBinoculars consistently maintains substantially higher mean F1 scores than Binoculars across all three generation models.

To provide a detailed overview across decoding configurations, Figure~\ref{fig:sampling_robustness} presents heatmaps of ROC AUC for both detectors under varying temperature and top-$p$ values. When temperature increases from 0.7 to 1.5 with top-$p$ fixed at 0.65, VietBinoculars consistently maintains high detection performance. However, its detection AUC drops substantially when temperature falls between 1.2 and 1.5 alongside top-$p$ between 0.80 and 1.0 for Sailor2-8B-Chat and Sailor2-14B-Chat. The third column of Figure~\ref{fig:sampling_robustness} displays the performance difference, defined as $\Delta\mathrm{AUC} = \mathrm{AUC}_{\mathrm{VietBinoculars}} - \mathrm{AUC}_{\mathrm{Binoculars}}$. These difference maps show positive values across most configurations, while the most pronounced negative differences appear at $T=1.5$ with top-$p\in\{0.95,1.0\}$.

These results directly extend the evaluation beyond greedy decoding, showing that strong discrimination persists across several sampled settings. However, detection performance degrades substantially under high temperatures and large top-$p$ values, which indicates that fixed thresholds do not provide uniform robustness across arbitrary decoding configurations. Consequently, the earlier near-perfect scores should be interpreted as benchmark-specific upper bounds rather than proof of general evasion resistance. While this experiment analyzes parameter variations during generation, Section~\ref{subsec:rewrite_robustness} investigates post-generation modifications using paraphrasing and rewriting techniques.

\subsection{Robustness to AI Text Rewriting}\label{subsec:rewrite_robustness}

To evaluate detector robustness against text manipulation, we examine three post-generation rewriting strategies, namely single paraphrasing, double paraphrasing, and human-style rewriting (simulated using targeted LLM prompt instructions). Sailor2-8B-Chat, Sailor2-14B-Chat, and Gemma-3-12B-it each rewrite their respective baseline continuations from the $(T,p)=(1.0,1.0)$ reference in Section~\ref{subsec:sampling_robustness}. The human passages remain unchanged across all evaluations. Because double paraphrasing relies on an independent intermediate output, the three rewritten variants\footnote{Detailed records and text samples for all rewritten datasets are available at \url{https://doi.org/10.57967/hf/10553}.} do not form a sequential pipeline.

\begin{figure}[!htb]
\centering
\includegraphics[width=\textwidth]{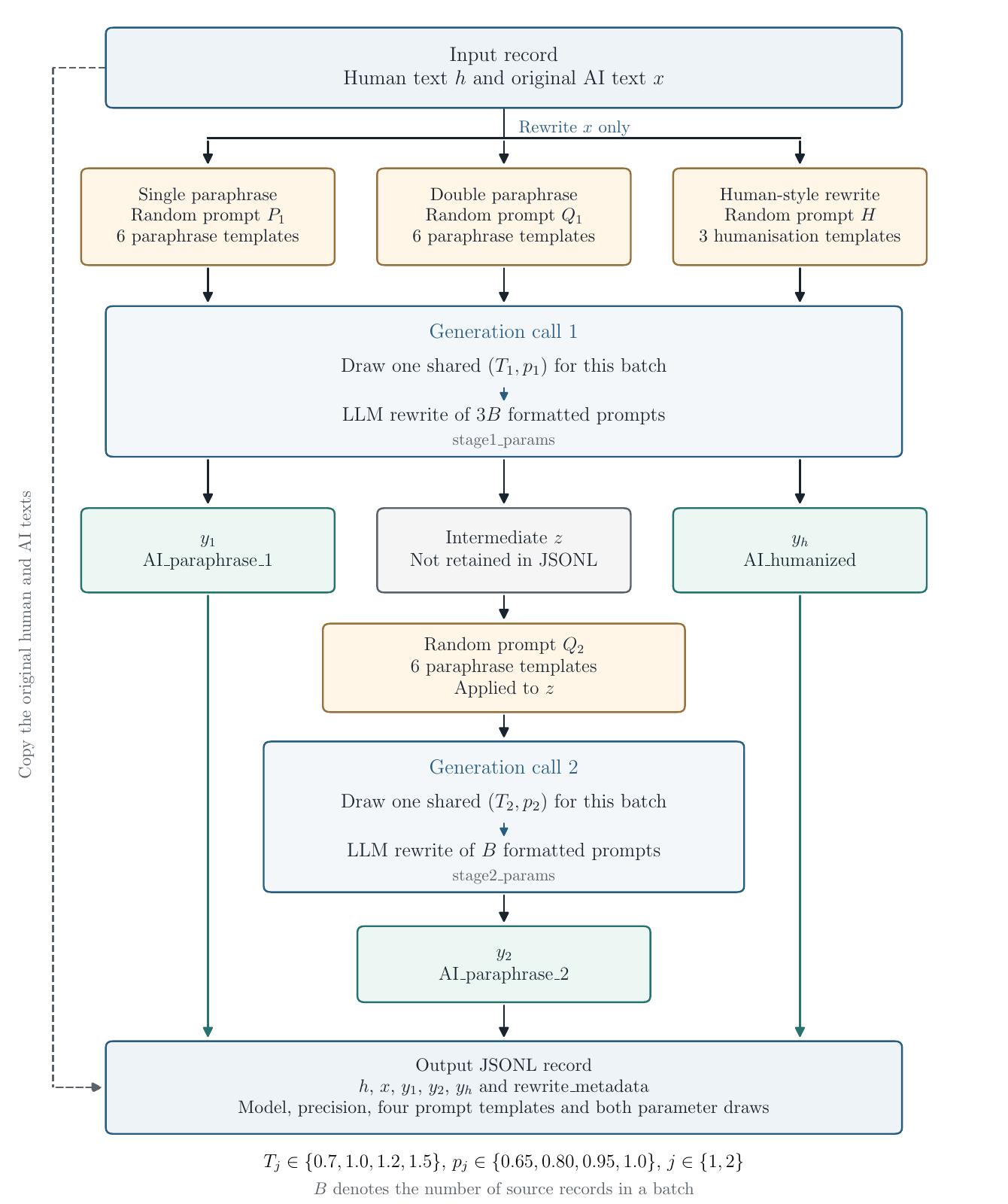}
\caption{Overview of the AI text rewriting pipeline. In the first stage, each batch of $B$ source records generates a single paraphrase, an intermediate text for double paraphrasing, and a human-style rewrite under a shared temperature and top-$p$ pair. The second stage subsequently rewrites these intermediate texts using new prompts and independently sampled parameters. The final output preserves the original texts alongside the three rewritten variants and their generation metadata.}
\label{fig:rewrite_pipeline}
\end{figure}

Each paraphrasing step randomly selects one of six prompt templates, while human-style rewriting selects one of three templates. All prompts are formulated in Vietnamese to approximate how Vietnamese-speaking users request text revisions. \ref{app:rewrite_prompts} provides the original templates alongside English translations included for reference. Temperature and top-$p$ values are drawn independently from $\{0.7,1.0,1.2,1.5\}$ and $\{0.65,0.80,0.95,1.0\}$. Each generation batch shares a common parameter pair, whereas the second paraphrasing step draws a new parameter configuration independently. All text transformations are evaluated without feedback from either detector. Figure~\ref{fig:rewrite_pipeline} summarizes this rewriting procedure.

\begin{figure}[!htb]
\centering
\includegraphics[width=\textwidth]{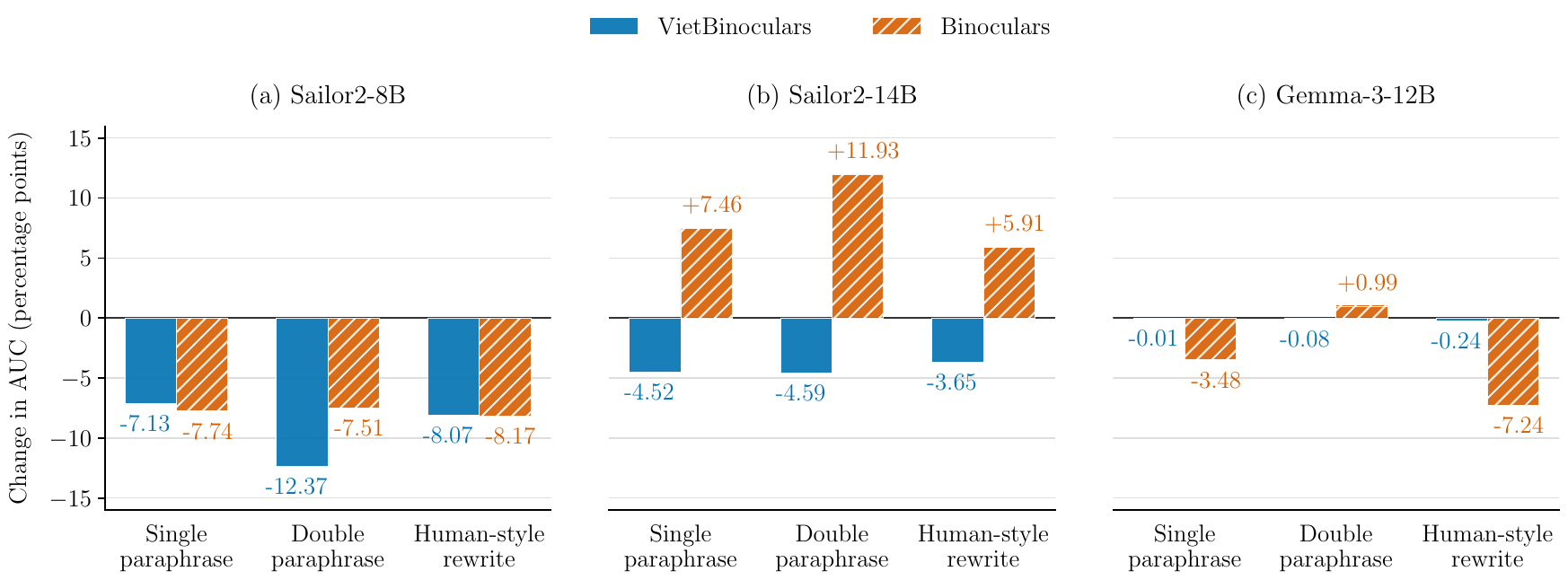}
\caption{Change in ROC AUC relative to the original AI texts, expressed in percentage points. Negative values indicate reduced discrimination, while positive values indicate increased discrimination. The three rewriting strategies are separate conditions. Bars show descriptive estimates without confidence intervals.}
\label{fig:rewrite_auc_change}
\end{figure}

\begin{table}[!htb]
\centering
\small
\setlength{\tabcolsep}{5pt}
\caption{ROC AUC of VietBinoculars and Binoculars on original and rewritten LLM-generated texts. Each variant is evaluated against human-written passages on a common subset across both detectors.  Values are rounded to three decimal places.}
\label{tab:rewrite_auc}
\begin{tabular}{llcc}
\toprule
Model & Text variant & \shortstack{AUC\\VietBinoculars} & \shortstack{AUC\\Binoculars} \\
\midrule
Sailor2-8B & Original AI text & 1.000 & 0.761 \\
 & Single paraphrase & 0.929 & 0.683 \\
 & Double paraphrase & 0.876 & 0.686 \\
 & Human-style rewrite & 0.919 & 0.679 \\
\midrule
Sailor2-14B & Original AI text & 0.952 & 0.851 \\
 & Single paraphrase & 0.907 & 0.925 \\
 & Double paraphrase & 0.906 & 0.970 \\
 & Human-style rewrite & 0.915 & 0.910 \\
\midrule
Gemma-3-12B & Original AI text & 1.000 & 0.788 \\
 & Single paraphrase & 1.000 & 0.753 \\
 & Double paraphrase & 0.999 & 0.798 \\
 & Human-style rewrite & 0.998 & 0.716 \\
\bottomrule
\end{tabular}
\end{table}
In order to evaluate detector performance across rewritten variants, ROC AUC is computed on matched pairs of human-written and LLM-generated texts. Table~\ref{tab:rewrite_auc} presents the absolute ROC AUC scores, while Figure~\ref{fig:rewrite_auc_change} illustrates the performance shift relative to the original AI texts in percentage points. 

The empirical results indicate that AI text rewriting impacts detection performance unevenly across generator architectures and rewriting strategies. For Gemma-3-12B, VietBinoculars demonstrates exceptional resilience, retaining ROC AUC scores above 0.998 across all rewritten variants. In contrast, detection capability on Sailor2-8B exhibits notable degradation, declining from an initial AUC of 1.000 to 0.929 for single paraphrasing, 0.919 for human-style rewriting, and 0.876 for double paraphrasing. Double paraphrasing induces the largest overall reduction for VietBinoculars, corresponding to a drop of 12.37 percentage points on Sailor2-8B (see Figure~\ref{fig:rewrite_auc_change} (a)). This vulnerability arises from the two-stage rewriting pipeline, where independent resampling of temperature and top-$p$ introduces cumulative lexical perturbations that shift token likelihoods away from original model signatures. A moderate decline also occurs on Sailor2-14B, where VietBinoculars drops between 3.65 and 4.59 percentage points. Conversely, baseline Binoculars displays an unexpected performance gain on Sailor2-14B, improving by 5.91 to 11.93 percentage points and reaching an AUC of 0.970 under double paraphrasing (see Figure~\ref{fig:rewrite_auc_change} (b)). Despite the observed degradation under iterative rewriting, VietBinoculars preserves an AUC exceeding 0.90 across all single paraphrased and human-style rewrites, confirming its practical robustness against standard rewriting attacks.

Overall, the rewriting experiments demonstrate that VietBinoculars retains robust zero-shot discrimination against single-pass paraphrasing and human-style modifications across all evaluated models. However, iterative transformations such as double paraphrasing degrade detection capability on the Sailor family, showing that multi-stage stochastic sampling can obscure generator-specific perplexity signatures. The stark contrast with Gemma-3-12B further highlights that resilience to text rewriting depends strongly on the underlying language model architecture. Together with the decoding parameter analysis in Section~\ref{subsec:sampling_robustness}, these findings demonstrate that while VietBinoculars withstands typical single-stage rewriting, its zero-shot resilience remains constrained by generator architecture and sampling depth.

\section{Discussion}\label{sec:discussion} 

On the original out-of-domain benchmarks with greedy decoding, VietBinoculars achieves AUROC exceeding 0.99 and detection accuracy of at least 98.78\% under the calibrated Youden's J threshold without retraining. Nevertheless, several methodological limitations of this initial setup should be recognized. First, we calibrated the optimal thresholds solely on the news domain using Sailor2-8B, while our literary evaluation relied exclusively on works by a single author, Vu Trong Phung. Consequently, these datasets may not fully capture the stylistic diversity and lexical breadth of Vietnamese text across different genres. Second, commercial detectors such as DetectGPT, GPTZero, and Turnitin were evaluated through black-box Web interfaces under default settings. Because internal prediction probabilities and calibration curves were inaccessible, we could not adjust these commercial baselines under identical decision thresholds. Third, zero-shot detection performance depends on sequence length. As demonstrated in our length analysis, passages shorter than 300 tokens provide fewer statistical signals, which leads to lower detection accuracy compared to longer texts.

The extended evaluations in Sections~\ref{subsec:sampling_robustness} and \ref{subsec:rewrite_robustness} further define the operational boundaries of the detector. The decoding grid demonstrates that while VietBinoculars retains high AUC for Gemma-3-12B-it, the Sailor models suffer substantial performance drops under high temperatures and top-$p$ values. Similarly, iterative rewriting reveals that multi-stage stochastic sampling can erode model-specific perplexity signatures, whereas single-pass modifications remain largely detectable. High AUC and effective classification at a fixed threshold also represent distinct operational outcomes. Consequently, the earlier near-perfect results should be interpreted as benchmark-specific upper bounds rather than evidence of universal resistance to evasion.

The VietBinoculars method also encounters deployment resource and inference efficiency challenges similar to those of the original Binoculars method. Although VietBinoculars does not require retraining or fine-tuning large language models for binary classification tasks, it still relies on two separate models, an observer and a performer, which increases computational and deployment overhead. Consequently, this effectively doubles the resource requirements for the detection system.

Real-world deployment further requires processing extensive documents that exceed the maximum context window of the PhoGPT-4B backbones. Rather than truncating lengthy submissions such as student essays or academic theses, an effective engineering strategy employs sliding-window chunking with configurable window and overlap sizes. Individual chunks are scored independently by VietBinoculars and document-level predictions are obtained through majority voting aggregation. This architecture has been operationalized in the open-source VietAIDetector system~\cite{nguyen2026vietaidetector}, which provides a web interface, OCR support for scanned documents, and downloadable verification reports.

Although VietBinoculars demonstrates robust performance across the evaluated settings, several practical evasion strategies remain unexamined, including English back-translation, manual text polishing, diacritic removal, and Vietnamese teencode substitution. These untested manipulations currently limit definitive conclusions regarding full-scale academic-integrity deployments. Because zero-shot statistical detectors remain susceptible to localized prediction errors and threshold sensitivity, automated scores should serve as preliminary screening rather than definitive judgments. Consequently, high-stakes decisions concerning academic misconduct or content authenticity must always be made under careful human supervision.

From a broader perspective, beyond natural language processing, zero-shot detection paradigms and statistical thresholding principles have also demonstrated potential across other specialized domains, such as target detection in synthetic aperture radar imagery \cite{zhang2025overview,he2025scattering}. Exploring connections between linguistic statistical signatures and generalized zero-shot detection across diverse signal modalities represents a promising direction for future research.

\section{Conclusion} \label{sec:conclusion} 

In this study, we introduced VietBinoculars, an effective zero-shot framework for detecting Vietnamese LLM-generated text without task-specific training. By pairing PhoGPT-4B as the observer and PhoGPT-4B-Chat as the performer, the method leverages dedicated Vietnamese BPE tokenization to prevent byte-level fragmentation and probability dilution. Our architectural ablation confirms that this compact 4B pair achieves superior detection sensitivity compared to massive multilingual alternatives such as Qwen2.5-7B. Across extensive out-of-domain benchmarks, VietBinoculars demonstrates exceptional discrimination, supported by paired McNemar significance tests ($p < 0.001$) and tight Wilson score confidence intervals, while also excelling on challenging Capybara prompts. Furthermore, extensive evaluations across 48 distinct model-decoding configurations and three post-generation rewriting strategies delineate the empirical boundaries of zero-shot detection. While VietBinoculars retains strong resilience against single-pass paraphrasing and human-style text revisions, detection sensitivity degrades notably under high-entropy sampling and iterative double paraphrasing.

For future research, we aim to extend evaluation to unexamined adversarial evasion attacks, including English back-translation, manual human polishing, diacritic removal, and Vietnamese teencode substitution. We also plan to investigate advanced aggregation methods for localized mixed-authorship detection, where human passages and AI generations are interwoven within a single document. Furthermore, we will explore adapting VietBinoculars to domain-specific tasks such as code generation detection in educational environments. Finally, utilizing VietBinoculars as an automated data-cleaning filter offers a promising direction to eliminate synthetic text from large-scale Vietnamese pre-training corpora.

\bibliographystyle{IEEEtranN}
\bibliography{ref}

\newpage
\appendix
\renewcommand{\thesection}{Appendix~\Alph{section}}
\renewcommand{\thesubsection}{\Alph{section}.\arabic{subsection}}
\section{Average Cross-Entropy Representation for log(\textit{perplexity})} \label{sec:appendixaveragecrossentropy}
Cross-entropy between two probability distributions $P$ and $Q$ over a set of events $X$ is defined as
\begin{equation}
H(P, Q) = -\sum_{x \in X} P(x) \log Q(x),
\end{equation}
where $P(x)$ is the probability of event $x$ according to distribution $P$, and $Q(x)$ is the probability of event $x$ according to distribution $Q$. Cross-entropy measures the difference between two probability distributions. In the context of $\log(\textit{perplexity})$, we consider cross-entropy as a measure of the difference between the empirical distribution $P$ of tokens in a sequence $s$ and the predicted distribution $P_{M_2}$ of tokens by the performer language model $M_2$. The empirical distribution $P$ is defined as 
\begin{equation}
P(v_j \vert x_{<i}) =\begin{cases} 
  1, \text{if } v_j = x_i, \\
  0, \text{otherwise}.
\end{cases} \label{eq:empirical_distribution}
\end{equation}
The cross-entropy between the empirical distribution $P$ and the predicted distribution $P_{M_2}$ at token $x_i$ in string $s$ is given by
\begin{equation}
H(x_i, P_{M_2}) = -\sum_{v_j \in \mathcal{V}} P(v_j \vert x_{<i}) \log P_{M_2}(v_j \vert x_{<i}). \label{eq:cross_entropy_perplexity}
\end{equation}
Substituting the empirical distribution in Eq.~\ref{eq:empirical_distribution} into Eq.~\ref{eq:cross_entropy_perplexity} at $v_j = x_i$ yields:
\begin{equation}
H(x_i, P_{M_2}) = -1\log P_{M_2}(x_i \vert x_{<i})=-\log\left(\mathcal{M}_2(s)_{i,x_i}\right). \label{eq:cross_entropy_perplexity_2}
\end{equation}
Based on Eq.~\ref{eq:cross_entropy_perplexity_2}, the average cross-entropy for the entire sequence $s$ with $L$ tokens can be expressed as
\begin{equation}
H(s, P_{M_2}) = \frac{1}{L} \sum_{i=1}^{L} H(x_i, P_{M_2}) = -\frac{1}{L} \sum_{i=1}^{L} \log\left(\mathcal{M}_2(s)_{i,x_i}\right).
\end{equation}
By combining the formulation from Eq.~\ref{eq:logppl_m2} with the above equation, the average cross-entropy can be rewritten as
\begin{equation}
H(s, \mathcal{M}_2) = \log{{\text{PPL}}_{\mathcal{M}_2}\left(s\right)}.
\end{equation}

\section{Illustration of Datasets} \label{sec:appendixdatasetillustration}
\subsection{Dataset structure}
The metadata example below describes the original dataset generation procedure, including greedy decoding. The additional sampled-generation configurations and the two retained greedy controls are documented separately in Section~\ref{subsec:sampling_robustness}.

There are two main keys in our datasets that should be noted.
The first key is ``text'', which corresponds to the human-written text samples.
The second key refers to text samples generated by LLMs, with dataset-specific naming conventions:
\begin{itemize}
  \item[$\star$] sail-Sailor2-8B-Chat-generated\_text for Sailor2-8B-OptiThreshold-News and Sailor2-8B-Validation-News Datasets,
  \item[$\star$] google-gemma-3-12b-it-generated\_text for Gemma-3-12B-News Dataset,
  \item[$\star$] sail-Sailor2-8B-Chat-hf\_generated\_text\_wo\_prompt for Sailor2-8B-VuTrongPhung Dataset,
  \item[$\star$] google-gemma-3-12b-it-hf\_generated\_text\_wo\_prompt for Gemma-3-12B-~VuTrongPhung Dataset.
\end{itemize}
In addition, our datasets also include a ``prompt'' key, , which corresponds to the prompt used by LLMs to generate the text samples. For example, in the Sailor2-8B-VuTrongPhung dataset, the prompt is a combination of ``\textit{system\_prompt}'' and ``\textit{user\_prompt}''. The ``prompt'' is defined as:
\begin{quote}
\vn{Bạn là một trợ lý văn học của tôi, hãy sinh ra một \{\textit{type}\} dựa vào đoạn văn dưới đây: $\backslash$n} \{\textit{user\_prompt}\}
\end{quote}
Its English translation is:
\begin{quote}
As my literary assistant, please generate a \{\textit{type}\} based on the following text: $\backslash$n \{\textit{user\_prompt}\}
\end{quote}
Here, \textit{type} refers to the category of Vu Trong Phung's writing we wish to generate, such as \textit{novel} or \textit{short story}. The ``\textit{user\_prompt}'' consists of the first 50 tokens of each chunk, as described in Section \ref{sec:vietnamese-dataset}. We also attach the LLM generation parameters under the ``gen\_meta'' key, which records the configuration used to produce the text samples. Finally, all datasets are stored in the \textit{.jsonl} format. The first line of the Sailor2-8B-OptiThreshold-News dataset is illustrated below:
\begin{quote}
\vn{
  \{
\texttt{"id"}: "bf3589f91e473f89",\\
\texttt{"text"}: "Theo thống kê của Sở Y tế Đắk Lắk, trong 6 tháng qua, toàn ngành y tế tỉnh có 36 viên chức y tế xin thôi việc. Trong đó, có 26 bác sĩ, 4 điều dưỡng, 1 nhân viên kỹ thuật và 5 viên chức ở các vị trí khác nhau. Trong số đó, Bệnh viện Đa khoa vùng Tây Nguyên là đơn vị có nhiều viên chức y tế thôi việc (16 bác sĩ). Mỗi trường hợp đều đưa ra lý do nghỉ việc khác nhau, tuy nhiên, theo ông Nay Phi La - Giám đốc Sở Y tế Đắk Lắk, nguyên nhân viên chức y tế xin thôi việc, bỏ việc xuất phát từ thực trạng quá tải công việc, ảnh hưởng của đại dịch COVID-19, môi trường làm việc áp lực và một số vị trí làm việc dễ bị lây nhiễm trong quá trình tiếp xúc bệnh nhân. Trong khi đó, chế độ tiền lương hưởng theo chức danh nghề nghiệp và các chế độ phụ cấp khác tính theo mức lương tối thiểu chung còn thấp, chưa tương xứng so với công việc, vị trí công tác. Ngoài ra, chính sách đãi ngộ đối với viên chức y tế đã có nhưng chưa tạo ra sự khác biệt lớn để khuyến khích cán bộ y tế yên tâm công tác tại các cơ sở y tế công lập dẫn đến tình trạng có nhiều viên chức, nhân viên y tế xin thôi việc, bỏ việc để chuyển công tác đến cơ sở y tế tư nhân, nơi có mức thu nhập cao hơn, ổn định hơn. Ông Nay Phi La thông tin thêm, địa phương đang đẩy mạnh tiến độ tiêm chủng vắc xin phòng COVID-19 và phòng, chống sốt xuất huyết đang diễn biến phức tạp, rất cần nhiều nhân lực. Do đó, loạt bác sĩ, nhân viên y tế nghỉ việc trong thời gian này đã phần nào ảnh hưởng đến công tác chăm sóc sức khỏe tại các cơ sở công lập. Thời gian tới, Sở Y tế tiếp tục tuyển dụng các vị trí còn thiếu trong chỉ tiêu biên chế; cấp tốc bồi dưỡng, đào tạo nguồn nhân lực tại chỗ. Bên cạnh đó, Sở Y tế Đắk Lắk đã có công văn kiến nghị Bộ Y tế đề xuất với Chính phủ xem xét ban hành chính sách tiền lương đặc thù cho đối tượng viên chức, nhân viên y tế; chính sách thu hút, ưu đãi, hỗ trợ đối với cán bộ y tế công tác ở lĩnh vực y tế dự phòng và tuyến y tế cơ sở nhằm tháo gỡ những khó khăn và thu hút viên chức ngành y tế yên tâm làm việc trong hệ thống y tế công lập.",\\
\texttt{"user\_prompt"}: "Theo thống kê của Sở Y tế Đắk Lắk, trong 6 tháng qua, toàn ngành y tế tỉnh có 36 viên chức y tế xin thôi việc. Trong đó, có 26 bác sĩ,",\\
\texttt{"topic"}: "unknown",\\
\texttt{"sail-Sailor2-8B-Chat-generated\_text"}: "4 điều dưỡng và 6 nhân viên khác.$\backslash$n$\backslash$nTóm tắt: Tỉnh Đắk Lắk đã chứng kiến sự gia tăng đáng kể về số lượng nhân viên y tế từ bỏ công việc của họ trong nửa đầu năm nay, với tổng cộng 36 người, bao gồm 26 bác sĩ, 4 điều dưỡng và 6 nhân viên hỗ trợ, theo dữ liệu được cung cấp bởi Sở Y tế địa phương. Tình hình này làm nổi bật những thách thức đang diễn ra mà hệ thống chăm sóc sức khỏe phải đối mặt ở khu vực này.$\backslash$n$\backslash$nCác yếu tố góp phần vào sự ra đi hàng loạt:$\backslash$n$\backslash$n1. **Lương thấp**: Mức lương cho các chuyên gia y tế thường không cạnh tranh so với chi phí sinh hoạt ngày càng tăng ở Đắk Lắk, dẫn đến cảm giác bất mãn tài chính giữa nhiều nhân viên.$\backslash$n$\backslash$n2. **Điều kiện làm việc kém**: Các cơ sở hạ tầng y tế, chẳng hạn như thiết bị lỗi thời, môi trường làm việc không an toàn và thiếu nguồn lực, ảnh hưởng tiêu cực đến tinh thần và hiệu suất của nhân viên.$\backslash$n$\backslash$n3. **Thiếu cơ hội phát triển nghề nghiệp**: Nhiều nhân viên y tế báo cáo rằng họ không thể tiếp cận đào tạo liên tục hoặc tiến bộ nghề nghiệp, khiến họ mất động lực và cam kết với vai trò hiện tại của mình.$\backslash$n$\backslash$n4. **Sự cân bằng giữa cuộc sống và công việc không phù hợp**: Với khối lượng công việc nặng nề và giờ làm việc kéo dài, nhân viên y tế đấu tranh để duy trì sự hài hòa giữa trách nhiệm cá nhân và chuyên môn của họ.$\backslash$n$\backslash$n5. **Hỗ trợ tâm lý xã hội hạn chế**: Thiếu dịch vụ tư vấn và hỗ trợ tâm lý cho nhân viên y tế, đặc biệt là sau khi đối phó với căng thẳng cao độ do COVID-19 gây ra, góp phần vào tỷ lệ nghỉ hưu cao hơn.$\backslash$n$\backslash$nNhững hậu quả tiềm năng:$\backslash$n$\backslash$nViệc mất mát nhân viên y tế quan trọng này đặt ra mối đe dọa nghiêm trọng đối với chất lượng dịch vụ chăm sóc sức khỏe, khả năng đáp ứng nhu cầu của bệnh nhân và sự ổn định tổng thể của hệ thống y tế. Để giải quyết tình trạng chảy máu chất xám này, cần thực hiện các biện pháp khẩn cấp, bao gồm:$\backslash$n$\backslash$n* **Tăng cường phúc lợi tài chính**: Xem xét",\\
\texttt{"gen\_meta"}: \{\\
    \texttt{"max\_new\_tokens"}: 512,\\
    \texttt{"do\_sample"}: false,\\
    \texttt{"pad\_token\_id"}: 151643,\\
    \texttt{"eos\_token\_id"}: 151645,\\
    \texttt{"use\_cache"}: true,\\
    \texttt{"num\_beams"}: 1\\
  \},\\
  \texttt{"model\_name"}: "sail$/$Sailor2-8B-Chat"\}
}
\end{quote}

\subsection{Token-length distributions per class} \label{subsec:token_length_distributions}
\begin{figure}[!htb]
  \centering
  \includegraphics[width=\textwidth]{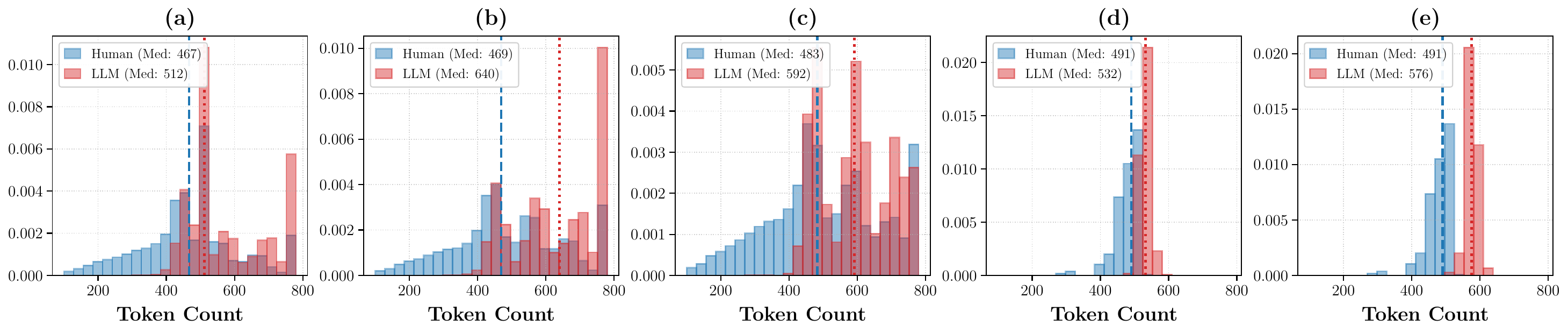}
  \caption{Token-length probability distributions per class across the five Vietnamese benchmarks, showing (a) Sailor2-8B-OptiThreshold-News, (b) Sailor2-8B-Validation-News, (c) Gemma-3-12B-News, (d) Sailor2-8B-VuTrongPhung, and (e) Gemma-3-12B-VuTrongPhung. Blue histograms and dashed vertical lines indicate human-written distributions and medians, respectively. Red histograms and dotted vertical lines denote LLM-generated distributions and medians, respectively.}
  \label{fig:token_length_distribution}
\end{figure}

Figure~\ref{fig:token_length_distribution} displays the empirical token-length probability distributions for human-written and LLM-generated texts across all five datasets. In the news domain, human articles exhibit a right-skewed distribution ranging from concise commentaries to extensive reporting, whereas LLM generations cluster more densely around their configured output budgets. In the literary domain, both human and LLM-generated texts concentrate tightly around 500 tokens. Although stylistic and lexical separation drives strong discrimination, the observed shift toward longer machine continuations highlights that length-correlated factors cannot be entirely excluded as potential contributors.

\subsection{Distribution of VietBinoculars scores with respect to sequence length on the Gemma-3-12B-News dataset}
\begin{figure}[!htb]
  \centering
  \includegraphics[width=0.9\textwidth]{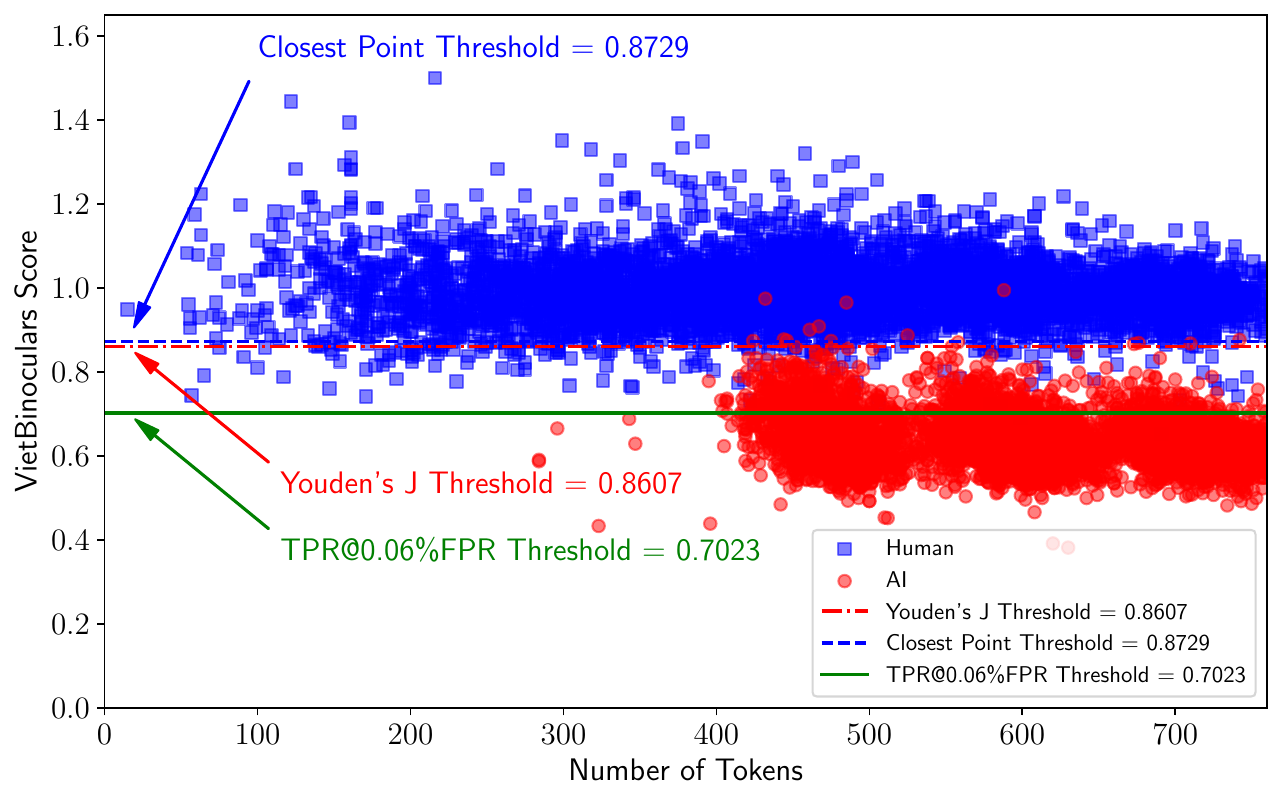}
  \caption{Distribution of VietBinoculars scores using BPE-encoded sequence length on the Gemma-3-12B-News dataset. The plotted lines include the classification thresholds: Youden's J, Closest Point, and TPR@0.06\%FPR. The y-axis indicates the VietBinoculars scores. Each point represents a text sample, with blue squares denoting human-written texts and red circles denoting LLM-generated texts.}
  \label{fig:binoculars_score_distribution}
\end{figure}

Figure~\ref{fig:binoculars_score_distribution} presents the distribution of VietBinoculars scores with respect to the number of tokens in the input texts from the Gemma-3-12B-News dataset. For visualization purposes, we also plot the three classification thresholds of VietBinoculars: Youden's J, Closest Point, and TPR@0.06\%FPR. Based on the distribution plot and the TPR@0.06\%FPR threshold, we can reasonably trust that texts with scores below this threshold are highly likely to be LLM-generated.

\section{Detailed Detection Results for the Capybara Problem in Vietnamese} \label{subsec:appendixcapybaradetectionresults} 
\begin{figure}[htbp]
  \centering
  \includegraphics[width=\textwidth]{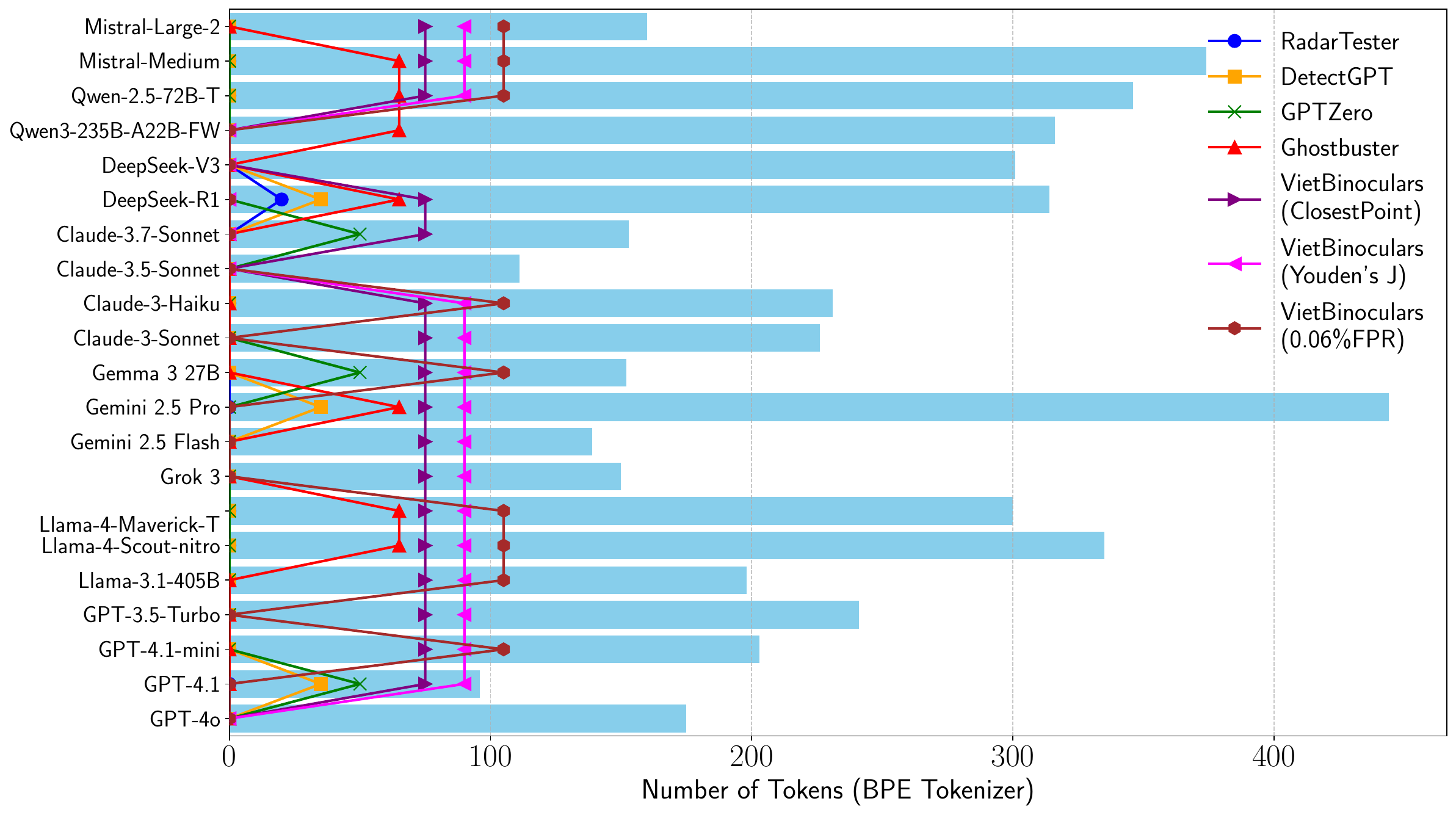}
  \caption{Detection accuracy of RadarTester, DetectGPT, GPTZero, Ghostbuster, and VietBinoculars on Vietnamese texts generated by various LLMs from the capybara prompt. The x-axis represents the number of tokens in text samples generated by LLMs using the capybara prompt, while the y-axis indicates the names of the LLMs used to generate the texts. The lines with markers show the detection accuracy of each method. If a detection method correctly identifies a text as LLM-generated, the corresponding marker is placed above zero; otherwise, the marker is positioned at zero.}
  \label{fig:capybara_detection_results}
\end{figure}

We employ the LLMs listed in Section~\ref{subsec:capybara_problem} to generate Vietnamese text from the capybara prompt. RadarTester, DetectGPT, GPTZero, and Ghostbuster are used as popular detection baselines for comparison with VietBinoculars. These detection methods are applied to identify Vietnamese LLM-generated texts produced by the aforementioned LLMs. We calculate the detection accuracy of each method using the optimal thresholds determined in Section~\ref{subsec:findglobaloptimalthreshold} for VietBinoculars, and the default threshold of 0.5 for the remaining methods. The detailed comparison results are presented in Figure~\ref{fig:capybara_detection_results}. Notably, all detection methods fail to identify texts generated by DeepSeek-V3, Claude-3.5-Sonnet, and GPT-4o as LLM-generated. VietBinoculars with the Closest Point threshold achieves the highest performance, with prediction failures occurring only for Qwen3-235B-A22B-FW, DeepSeek-V3, Claude-3.5-Sonnet, and GPT-4o.

\section{Inference latency} \label{sec:appendix_runtime_benchmark}

We evaluated the inference runtime on the Sailor2-8B-VuTrongPhung literary dataset using 180 text pairs. All measurements were synchronized on the GPU to ensure timing accuracy. Two initial warmup batches were executed to eliminate CUDA kernel initialization overhead.

\begin{figure}[!htb]
  \centering
  \includegraphics[width=0.85\textwidth]{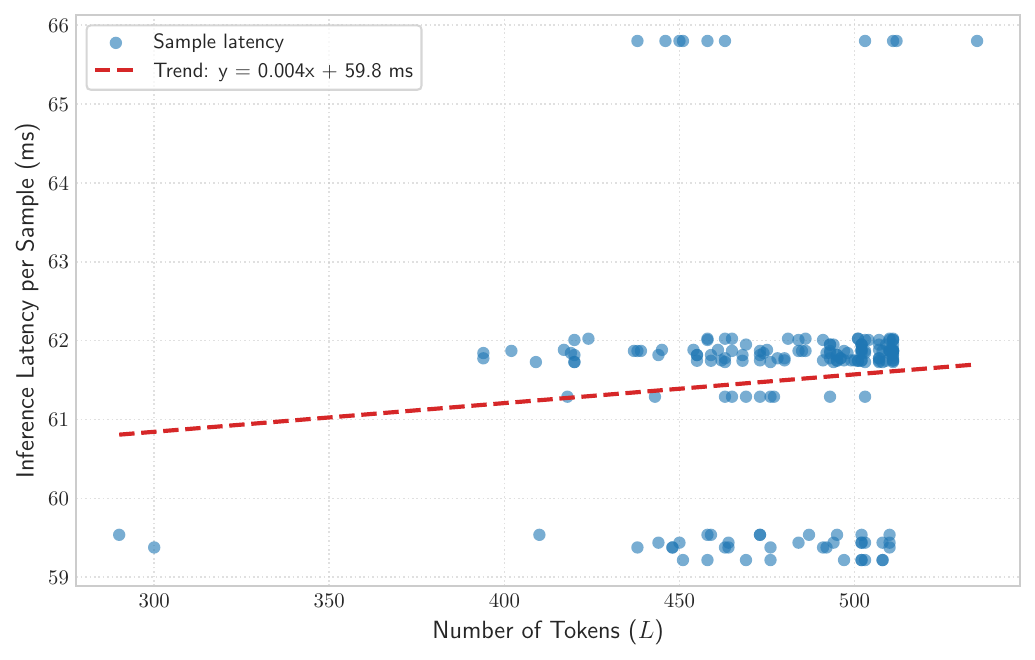}
  \caption{Inference latency per sample of VietBinoculars across different token lengths on the Vietnamese literary dataset. Measurements reflect synchronized GPU execution after two warmup batches. The dashed line depicts the linear trend line.}
  \label{fig:runtime_vs_tokens}
\end{figure}

Inference latency exhibits high consistency across evaluated samples. The median latency is 61.80 ms per sample, with a standard deviation of 1.46 ms and a range between 59.22 ms and 65.80 ms. Figure~\ref{fig:runtime_vs_tokens} presents the latency per sample across different sequence lengths. The fitted linear trend line shows a negligible slope of 0.004 ms per token. This confirms that per-sample latency is largely independent of text length within this range, because the GPU executes padded batches in parallel.

\section{Vietnamese Prompt Templates for Text Rewriting}\label{app:rewrite_prompts}

Table~\ref{tab:rewrite_prompts} presents the exact prompt templates used in Section~\ref{subsec:rewrite_robustness}, accompanied by English translations for reference. Only the original Vietnamese prompts were employed during text generation, preserving their exact punctuation and line breaks. The placeholder \texttt{\{\}} indicates where the input text is inserted, which corresponds to the intermediate generated output in the second step of double paraphrasing. Template identifiers denote catalog entries and remain distinct from the schematic prompt symbols shown in Figure~\ref{fig:rewrite_pipeline}.

\begingroup
\small
\setlength{\tabcolsep}{3pt}
\renewcommand{\arraystretch}{1.12}
\begin{longtable}{@{}p{0.10\textwidth}p{0.15\textwidth}>{\raggedright\arraybackslash}p{0.35\textwidth}>{\raggedright\arraybackslash}p{0.35\textwidth}@{}}
\caption{Original Vietnamese rewriting prompts and English translations. The translations were not used to generate the evaluated texts.}\label{tab:rewrite_prompts}\\
\toprule
Template ID & Rewriting strategy & Original Vietnamese prompt & English translation \\
\midrule
\endfirsthead
\multicolumn{4}{l}{Table \thetable. Continued}\\
\toprule
Template ID & Rewriting strategy & Original Vietnamese prompt & English translation \\
\midrule
\endhead
\midrule
\multicolumn{4}{r}{Continued on the next page}\\
\endfoot
\bottomrule
\endlastfoot
P1 & Paraphrasing & {\vn Hãy viết lại đoạn văn sau với văn phong tự nhiên hơn, giữ nguyên ý nghĩa:\newline{}\mbox{}\newline{}\{\}} & {\en Rewrite the following passage in a more natural style while preserving its meaning:\newline{}\mbox{}\newline{}\{\}} \\[5pt]
P2 & Paraphrasing & {\vn Diễn đạt lại nội dung dưới đây bằng cách khác nhưng không thay đổi ý nghĩa:\newline{}\mbox{}\newline{}\{\}} & {\en Express the content below differently without changing its meaning:\newline{}\mbox{}\newline{}\{\}} \\[5pt]
P3 & Paraphrasing & {\vn Viết lại đoạn văn sau theo cách dễ hiểu và tự nhiên hơn:\newline{}\mbox{}\newline{}\{\}} & {\en Rewrite the following passage in a clearer and more natural way:\newline{}\mbox{}\{\}} \\[5pt]
P4 & Paraphrasing & {\vn Hãy paraphrase đoạn văn sau:\newline{}\mbox{}\newline{}\{\}} & {\en Paraphrase the following passage:\newline{}\mbox{}\{\}} \\[5pt]
P5 & Paraphrasing & {\vn Rewrite đoạn sau với cách diễn đạt khác nhưng vẫn giữ nguyên ý nghĩa:\newline{}\mbox{}\newline{}\{\}} & {\en Rewrite the following passage using different wording while preserving its meaning:\newline{}\mbox{}\newline{}\{\}} \\[5pt]
P6 & Paraphrasing & {\vn Paraphrase đoạn văn sau sao cho tự nhiên hơn:\newline{}\mbox{}\newline{}\{\}} & {\en Paraphrase the following passage to make it sound more natural:\newline{}\mbox{}\{\}} \\[5pt]
H1 & Human-style rewriting & {\vn Hãy viết lại đoạn văn sau sao cho giống văn bản do con người viết:\newline{}- Có thể thêm cảm xúc nhẹ\newline{}- Có thể thay đổi cấu trúc câu\newline{}- Tránh quá trôi chảy như AI\newline{}\mbox{}\newline{}Đoạn văn:\newline{}\{\}} & {\en Rewrite the following passage so that it resembles text written by a human:\newline{}- You may add subtle emotion\newline{}- You may change the sentence structure\newline{}- Avoid excessive fluency like that of AI\newline{}\mbox{}\newline{}Passage:\newline{}\{\}} \\[5pt]
H2 & Human-style rewriting & {\vn Viết lại đoạn sau theo phong cách tự nhiên như người viết thật:\newline{}- Không cần quá hoàn hảo\newline{}- Có thể thay đổi cách diễn đạt\newline{}\mbox{}\newline{}\{\}} & {\en Rewrite the following passage in a natural style, as if written by a real person:\newline{}- It does not need to be perfect\newline{}- You may change the wording\newline{}\mbox{}\newline{}\{\}} \\[5pt]
H3 & Human-style rewriting & {\vn Chỉnh sửa đoạn văn sau để giống văn phong con người hơn, tránh quá máy móc:\newline{}\mbox{}\newline{}\{\}} & {\en Edit the following passage to make its style more human-like and less mechanical:\newline{}\mbox{}\newline{}\{\}} \\[5pt]
\end{longtable}
\endgroup

\section{Ablation Study on Detector Model Pairs} \label{sec:appendix_model_pair_ablation}

To validate whether PhoGPT-4B represents an optimal architectural choice rather than an arbitrary selection, we conduct an ablation study comparing alternative model pairs across our out-of-domain benchmarks. Specifically, we evaluate a multilingual pair consisting of Qwen2.5-7B as the observer and Qwen2.5-7B-Instruct as the performer. We select Qwen2.5-7B rather than Sailor2-8B to prevent self-detection bias, because Sailor2-8B serves as a text generator in our benchmark suite. Evaluating a generator with its own architectural family would introduce circular white-box exposure artifacts, whereas Qwen2.5-7B provides a strictly independent evaluation. In addition, Qwen2.5 represents a competitive state-of-the-art multilingual model with robust Vietnamese capabilities. Table~\ref{tab:model_pair_ablation} summarizes the detection AUROC across the evaluated configurations.
\begin{table}[!ht]
\centering
\small
\caption{Ablation of detector model pairs across Vietnamese out-of-domain benchmarks (ROC AUC).}
\label{tab:model_pair_ablation}
\resizebox{\textwidth}{!}{
\begin{tabular}{lcccc}
\toprule
\textbf{Model Pair} & \textbf{Model Size} & \shortstack{\textbf{Gemma-3-12B-}\\\textbf{News}} & \shortstack{\textbf{Gemma-3-12B-}\\\textbf{VuTrongPhung}} & \shortstack{\textbf{Sailor2-8B-}\\\textbf{VuTrongPhung}} \\
\midrule
Falcon-7B (Original Binoculars) & 2 $\times$ 7B & 0.784 & 0.789 & 0.761 \\
Qwen2.5-7B (Multilingual Pair) & 2 $\times$ 7B & 0.969 & 1.000 & 0.994 \\
PhoGPT-4B (VietBinoculars) & 2 $\times$ 4B & \textbf{0.999} & \textbf{1.000} & \textbf{1.000} \\
\bottomrule
\end{tabular}}
\end{table}
These empirical findings demonstrate that PhoGPT-4B is the best-performing overall among the evaluated detector pairs on these benchmarks.

First, dedicated Vietnamese tokenization provides superior scoring sensitivity. Although the multilingual Qwen2.5-7B pair achieves strong discriminative performance and substantially outperforms English-centric Falcon-7B, PhoGPT-4B consistently yields the highest overall scores above 0.999. Its specialized Vietnamese BPE vocabulary avoids the token fragmentation and probability dilution typical of massive multilingual vocabularies.

Second, PhoGPT-4B offers superior computational and deployment efficiency. It requires only 4 billion parameters per model, which enables practical dual-model deployment on a single commodity GPU with low inference latency. In contrast, 7B pairs require 14 billion combined parameters and double the memory footprint without delivering higher detection accuracy.

\end{document}